\documentclass{article}
\makeatletter
\def\input@path{{src/}}
\makeatother
\usepackage[preprint]{iclr2027_conference}
\usepackage[OT1]{fontenc}
\usepackage{times}

\usepackage{amsmath,amsfonts,bm}

\def\eqref#1{equation~\ref{#1}}

\def\1{\bm{1}}

\DeclareMathAlphabet{\mathsfit}{\encodingdefault}{\sfdefault}{m}{sl}
\SetMathAlphabet{\mathsfit}{bold}{\encodingdefault}{\sfdefault}{bx}{n}

\usepackage{hyperref}
\usepackage{url}
\usepackage{amsmath,amssymb,amsfonts}
\usepackage{algorithm}
\usepackage{algorithmic}
\usepackage{graphicx}
\usepackage{booktabs}
\usepackage{multirow}
\usepackage{array}
\usepackage{tabularx}
\usepackage{threeparttable}
\usepackage{wrapfig}
\usepackage{needspace}
\usepackage{fvextra}
\usepackage{xcolor}
\usepackage{xspace}
\usepackage{tikz}
\usetikzlibrary{arrows,positioning,fit,calc,plotmarks}

\graphicspath{{figs/}}

\newcommand{\gain}[1]{\textcolor{red}{\(\uparrow\)#1}}
\newcommand{\drop}[1]{\textcolor{green!50!black}{\(\downarrow\)#1}}
\newcommand{\ours}{{\textsc{TreeFuse}}\xspace}
\newcommand{\scoreci}[2]{#1\,{\scriptsize\(\pm\)#2}}

\title{Rethinking Retrieval in the Long-Context Era: Training-Free Semantic Trees for Adaptive Context Construction}

\author{Zhe Xu$^{1,}$\thanks{These authors contributed equally.}, Tianhao Liao$^{1,*}$, Zhenqing Ling$^{2,*}$, Ying Shen$^{1}$, Yaliang Li$^{2}$, Daoyuan Chen$^{2}$ \\[0.5em]
\normalfont $^{1}$Sun Yat-sen University \quad $^{2}$Alibaba Group
}

\hypersetup{
  pdftitle={Rethinking Retrieval in the Long-Context Era: Training-Free Semantic Trees for Adaptive Context Construction},
  pdfauthor={Zhe Xu, Tianhao Liao, Zhenqing Ling, Ying Shen, Yaliang Li, Daoyuan Chen}
}

\begin{document}
\maketitle

\begin{abstract}
Long-context language models can increasingly process entire documents, making
full-context prompting a practical alternative to passage retrieval. However,
greater capacity determines how much information a model can receive, but not
which information best supports a query. This distinction motivates rethinking
retrieval: when the full document already fits, can selecting and organizing
its evidence improve answering? The challenge is to retain supporting facts
and relations while keeping the context focused on the question. To reconcile
these goals, we introduce \ours, a training-free framework that uses semantic
abstraction to guide selection while retaining original text for answering.
\ours constructs a reusable semantic hierarchy from the full document in a
single generation, linking node descriptions to source spans. Given a query,
structural--semantic scoring and fine-grained lexical attribution nominate
shared source nodes; their union with ancestor context preserves structural
relations and complete local text. Across four frontier answer models, \ours
improves average QA-F1 by 1.72 points over full-context prompting on three
LongBench tasks, surpassing the evaluated baselines on the four-model average
while reducing answer-context length by 58.55\%. Our code is available at the \href{https://anonymous.4open.science/r/treefuse-ECFE}{code repository}.
\end{abstract}
\section{Introduction}
\label{sec:intro}

Frontier language models can increasingly accommodate entire documents within
their context windows, making full-context prompting a practical option for
long-document question answering. Yet input capacity and context selection
address different questions: how much information a model can receive, and
which information it should receive for a particular query. Prior work shows
that performance can depend strongly on where relevant information appears
within a long context, even when that information remains available to the
model \citep{liu2024lost}. Prior compression methods show that selecting context
can improve answering as well as shorten the input
\citep{longllmlingua2024,provence2025}, while LaRA \citep{lara2025} finds no
universally preferable choice between retrieval-based and full-context
approaches. These findings motivate rethinking retrieval beyond accommodating
input-length constraints: \textbf{When the full document already fits within
the context window, can retrieval improve answering, and how should it select
the supporting context?}

A relevant passage is not always sufficient to support an answer. A question
may require a directly mentioned fact together with background or a relation
established elsewhere in the document. Sufficient Context
\citep{joren2025sufficient} explicitly distinguishes contextual support from
answer correctness. Context construction must therefore consider both the
support it retains and how selectively it presents that support. Narrow
extraction can omit a necessary fact or separate a detail from its
interpretation, whereas broad expansion may introduce unrelated material.
Controlled studies of arithmetic reasoning show that such irrelevant
information can interfere with model predictions \citep{shi2023distracted},
motivating attention to context concentration as well as support retention.
Hierarchical methods such as HiChunk \citep{hichunk2025} and LongRefiner
\citep{longrefiner2025} make document organization available to retrieval and
refinement. The remaining design question is how to use that organization while
remaining sensitive to precise source-level details.

\begin{wrapfigure}{r}{0.48\textwidth}
\setlength{\abovecaptionskip}{-12pt} 
\setlength{\belowcaptionskip}{0pt}
\centering
\includegraphics[width=\linewidth]{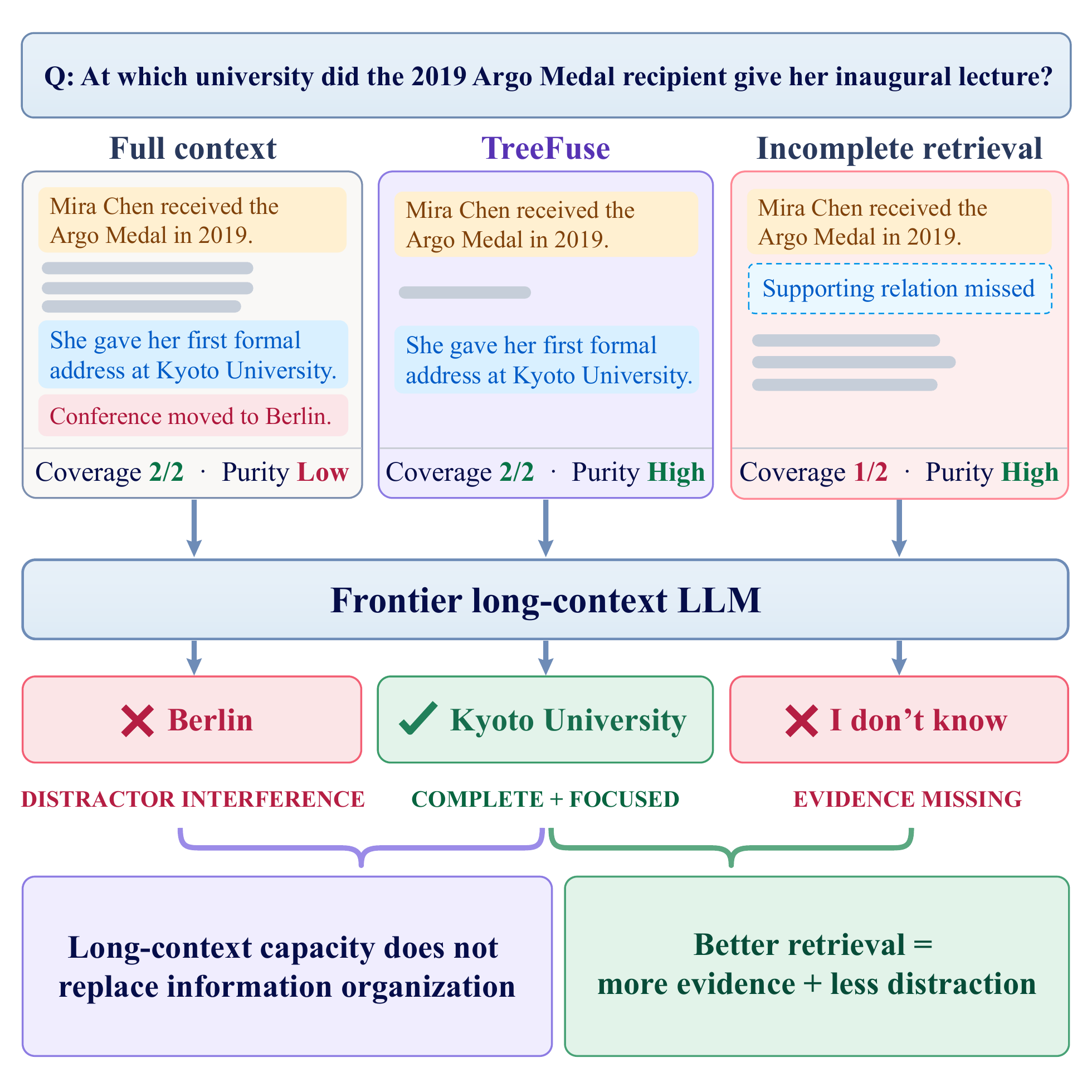}
\caption{
A schematic example contrasts full context with distractors, incomplete retrieval with
missing support, and \ours's focused context retaining both supporting statements.}
\label{fig:example}
\vspace{-\baselineskip}
\end{wrapfigure}


To reconcile support retention with context concentration, we separate the
information used to select evidence from the text supplied for answering.
Figure~\ref{fig:example} illustrates the goal: retaining supporting
facts and their relation while excluding unrelated material. We instantiate
this principle with \ours, a training-free framework that uses semantic
abstraction to guide selection while preserving original evidence.
\textbf{Step 1: Source-grounded Semantic Tree Construction.} A pretrained model
jointly proposes semantic boundaries and hierarchical relations from the full
document in a single generation, linking node descriptions to original source
spans. The resulting hierarchy is query-independent and reusable.
\textbf{Step 2: Query-adaptive Context Construction.} Structural--semantic
scoring and fine-grained lexical attribution nominate nodes from this shared
representation. A localized lexical match can retain a complete source node,
supplementing semantic judgments with precise details. The selected nodes and
their ancestors form a pruned hierarchy that preserves the original
parent--child relations and complete local text. The same constructed context
can be supplied to different answer models without model-specific fine-tuning.

We evaluate \ours with four frontier answer models that offer strong
long-context processing capabilities: Qwen3.8-Max \citep{qwen38max2026},
DeepSeek-V4-Pro \citep{deepseekv4doc2026}, Kimi-K2.6 \citep{kimik26blog2026},
and GLM-5.3 \citep{glm53docs2026}. Full-context prompting provides a direct
reference for testing whether retrieval remains beneficial when the complete
document is already available to the answer model. Across three LongBench QA
tasks, \ours reduces the answer-context length by 58.55\% on average while
improving the four-model average by 1.72 QA-F1 points over full-context prompting.
The gains are more pronounced on the multi-hop HotpotQA and 2WikiMQA tasks,
where the four-model averages improve by 1.72 and 2.46 points, respectively.
On 2WikiMQA with Kimi-K2.6, \ours achieves a 7.00-point improvement,
corresponding to a 14.76\% relative gain. Evaluation on LaRA further tests
transfer across heterogeneous long-document tasks, where \ours achieves the
highest four-model average accuracy among the evaluated methods with 43.87\%
less answer context. Beyond answer scores, source-evidence analyses examine
completeness and concentration, while component comparisons characterize
semantic and lexical selection, hierarchical organization, and single-pass
construction. These evaluations address both parts of our research question,
establishing average gains over full-context prompting and examining how the
selected contexts retain and organize supporting evidence.

To summarize, our contributions are:
\begingroup
\setlength{\leftmargini}{1em}
\setlength{\labelsep}{0.4em}
\begin{itemize}
\item 
We empirically demonstrate
that query-adaptive context construction can improve average answering
performance even when full-document input is feasible, positioning retrieval
as a complement to long-context capacity.
\item 
We introduce reusable
semantic trees and dual-granularity selection to coordinate structural support
with precise source details, retaining complete local text without
method-specific training.
\item We provide a broad evaluation across four frontier answer-model families
and diverse long-document tasks, obtaining a 1.72-point average QA-F1 improvement on the three
LongBench tasks with 58.55\% less context. We release our code to support
research on adaptive context construction: \url{https://anonymous.4open.science/r/treefuse-ECFE}.
\end{itemize}
\endgroup
\section{Related Work}
\label{sec:related}

\paragraph{Retrieval and Compression for Long-Context Question Answering}
LongBench \citep{longbench2024} and LaRA \citep{lara2025} evaluate long-context
understanding and retrieval-based answering. LongRAG by \citet{jiang2024longrag}
groups related documents into larger retrieval units to exploit long-context
models. A separate LongRAG by \citet{zhao2024longrag} jointly addresses global
information and factual details in long-context QA. Compression methods act at
different granularities: LongLLMLingua \citep{longllmlingua2024} prunes tokens,
Provence \citep{provence2025} selects sentences, and RECOMP \citep{recomp2024}
and BRIEF-Pro \citep{briefpro2026} learn or generate compact contexts. ACC-RAG
\citep{accrag2025} adapts compression rates, while SARA \citep{sara2025} combines
textual and compressed representations. Sufficient Context
\citep{joren2025sufficient} examines whether supplied information is sufficient
for answering. 

\paragraph{Semantic Chunking and Hierarchical Document Indexing}
Semantic chunkers such as LumberChunker \citep{lumberchunker2024} and
Meta-Chunking \citep{metachunking2025} seek coherent boundaries, with libraries
such as Chonkie\footnote{\url{https://docs.chonkie.ai/}} providing multiple
implementations. Hierarchical indexes additionally organize relationships among
units. RAPTOR \citep{raptor2024} recursively clusters and summarizes passages.
TreeRAG \citep{tao2025treerag} combines tree-based chunking with bidirectional
traversal. SiReRAG \citep{zhang2025sirerag} constructs similarity and relatedness
trees and combines their nodes in a retrieval pool. Heading-aware indexing
\citep{headingaware2025} exploits existing sections, while HiChunk
\citep{hichunk2025} and LongRefiner \citep{longrefiner2025} learn specialized
components for hierarchical chunking or refinement. 

\paragraph{Structure-Aware and Hybrid Retrieval}
BM25 \citep{robertson2009bm25} and DPR \citep{karpukhin2020dpr} provide lexical
and learned semantic relevance signals. Contextual Retrieval
\citep{anthropic2024contextual} prepends explanatory context to chunks before
embedding and BM25 indexing. MemWalker \citep{chen2023memwalker} interactively
navigates summary trees, while PageIndex \citep{pageindex2025} uses
reasoning-based search over document hierarchies. HiChunk \citep{hichunk2025}
expands retrieval hits through Auto-Merge, and LongRefiner \citep{longrefiner2025}
performs query-aware refinement. DISRetrieval \citep{chen2026disretrieval}
combines discourse parsing, LLM-enhanced node representations, and
structure-guided hierarchical retrieval. 

\ours approaches retrieval as a complement to long-context capacity, not solely
a remedy for limited input length. It targets answer contexts that retain
supporting evidence while concentrating on the query, using a reusable semantic
hierarchy to guide source-preserving selection. Appendix~\ref{app:related-comparison}
provides a detailed comparison with evaluated baselines and selected related methods.

\section{Method}
\label{sec:method}

\subsection{Problem Formulation and Overview}
Given a document $D$ and query $q$, \ours constructs an answer context $C(q,D)$
by selecting source-grounded semantic units. The objective is to retain
supporting information while limiting content unrelated to the query, even when
the full document fits within the model's context window.

Source selection does not create new source evidence; it changes which content
is visible to the answer model and how that content is organized. \ours
therefore separates the information used to select evidence from the text
delivered for answering. A compact semantic view exposes a node's topic and
structural role, while its original text retains details not captured by the
description. Selecting complete local units preserves the surroundings of a
match, and ancestor completion retains structural connections without including
every neighboring branch.

\begin{figure}[!t]
\centering

\includegraphics[width=\textwidth]{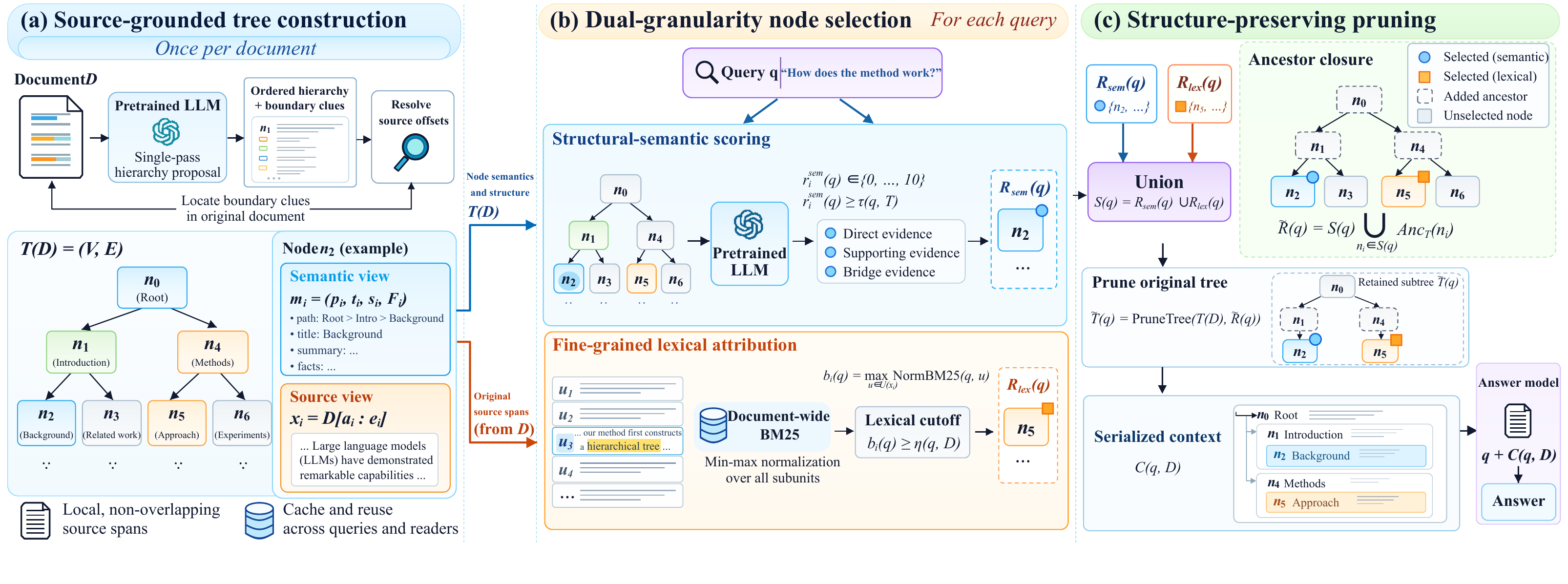}
\vspace{-0.3in}
\setlength{\belowcaptionskip}{0pt}
\caption{Overview of TreeFuse. (a) A single generation constructs a semantic hierarchy whose nodes are aligned to original source spans. (b) Structural–semantic scoring and fine-grained lexical attribution independently nominate nodes from this shared representation. (c) Their union is completed with ancestor nodes and used to prune the original tree. Retained nodes preserve their structural relations and complete local source text for answering.}
\label{fig:method-overview}
\vspace{-\baselineskip}
\end{figure}

\ours combines a reusable hierarchy with query-dependent selection and pruning.
Figure~\ref{fig:method-overview} outlines the pipeline. First, source-grounded semantic tree construction links node
descriptions and hierarchical relations to original source spans
(\S\ref{sec:semantic-tree}). Given a query, structural--semantic scoring and
fine-grained lexical attribution nominate nodes from this shared representation
(\S\ref{sec:dual-selection}). Their union is completed with ancestor nodes and
used to prune the original tree, yielding an answer context that preserves
retained structural relations and complete local source text
(\S\ref{sec:context-assembly}).

\subsection{Source-Grounded Semantic Tree Construction}
\label{sec:semantic-tree}
The document-to-hierarchy mapping in Figure~\ref{fig:method-overview} links
document structure to original source text. A pretrained language model receives
the full document and jointly proposes semantic boundaries and parent--child
relations in one main generation. Its output contains an ordered hierarchy,
node titles, summaries, atomic facts, and verbatim boundary clues. A
source-alignment procedure resolves these clues to document offsets, yielding
a hierarchy $T(D)=(V,E)$.

Each node $n_i\in V$ has a semantic view $m_i$ and a source view $x_i$:
\begin{equation}
    m_i=(p_i,t_i,s_i,F_i),\qquad x_i=D[a_i:e_i],
\end{equation}

where $p_i$ is its structural path, $t_i$ its title, $s_i$ its summary, $F_i$
its atomic facts, and $[a_i,e_i)$ its source interval. Each node owns a local
source span rather than the concatenation of its descendants. Its identity
links the semantic view, source interval, and position in the hierarchy, so
semantic scoring and lexical matching can nominate the same underlying source
unit. The descriptions provide an indexing view; the original spans provide
answer evidence. This source-grounded hierarchy is constructed independently
of queries and can be reused across queries and answer models. Construction
prompts, source alignment, and a case illustrating the tree's utility for
downstream QA appear in Appendices~\ref{app:prompt-templates},
\ref{app:tree-construction}, and~\ref{app:boundary-adequacy-case}, respectively.

\subsection{Query-Adaptive Dual-Granularity Selection}
\label{sec:dual-selection}
The two channels in the dual-granularity selection module of
Figure~\ref{fig:method-overview} nominate nodes from the shared hierarchy using
complementary views. Structural--semantic scoring assesses a node's relevance
and supporting role from its descriptions and structural context. Fine-grained
lexical attribution identifies localized matches in its original text, including
entities, values, or phrases not captured by the descriptions. The channels
produce node sets independently rather than averaging scores with different
scales and meanings. Their selections may overlap: a node can be nominated by
either channel or both.

\paragraph{Structural--semantic scoring.}
For each query, a pretrained model rates every node on a $0$--$10$ relevance
scale using its semantic view, available parent metadata, and document-level
structural context. The prompt requests judgments of direct relevance and
supporting relationships. All nodes are scored, so a child remains eligible
even if its parent scores low. The initial selection is
\[
R_{\mathrm{sem}}^{0}(q)=
\{n_i\in V:r_i^{\mathrm{sem}}(q)\geq\tau(q,T)\},
\]
where $r_i^{\mathrm{sem}}(q)$ is the node's relevance rating and $\tau(q,T)$
the semantic cutoff. A nonempty-selection safeguard yields the final set
$R_{\mathrm{sem}}(q)$. Appendix~\ref{app:selection-thresholds} details the cutoff
policy and safeguards.

\paragraph{Fine-grained lexical attribution.}
Each source span $x_i$ is divided into sentence- or paragraph-like subunits
$\mathcal U(n_i)$. We build a shared BM25 index over the document-wide pool
$\mathcal U=\bigcup_i\mathcal U(n_i)$, min--max normalize scores across this
pool, and attribute the strongest match to its enclosing node:
\begin{equation}
    b_i(q)=\max_{u\in\mathcal U(n_i)}\operatorname{NormBM25}(q,u),
    \qquad
    R_{\mathrm{lex}}(q)=\{n_i\in V:b_i(q)\geq\eta(q,D)\}.
\end{equation}

A node without subunits receives a zero lexical score. A localized match can
therefore nominate a complete node without requiring its entire source span to
have high lexical overlap. The lexical attribution channel in
Figure~\ref{fig:method-overview} maps fine-grained matches to their enclosing
source nodes: the subunit determines the matching granularity, while the node
determines the retention granularity. The lexical cutoff $\eta(q,D)$ follows
the P90 rule in Appendix~\ref{app:selection-thresholds}, which also specifies
score normalization.

\subsection{Structure-Preserving Pruning and Context Construction}
\label{sec:context-assembly}
The union and ancestor-closure step in Figure~\ref{fig:method-overview}
determines the node set for a query-specific pruned tree. We first take the
union of the two candidate sets:
\begin{equation}
    S(q)=R_{\mathrm{sem}}(q)\cup R_{\mathrm{lex}}(q).
\end{equation}

Lexical matching supplements semantic selection rather than filtering it, so
a node nominated by either channel remains eligible for retention. We then
add the ancestors of the selected nodes:
\begin{equation}
    \widetilde{R}(q)=S(q)\cup
    \bigcup_{n_i\in S(q)}\operatorname{Anc}_T(n_i).
\end{equation}

Ancestor completion preserves the paths connecting retained nodes within the
original hierarchy. It does not expand a selected node to every surrounding
branch: each added ancestor contributes only its own local source span, not
the text of all its descendants. We prune the original tree to this
ancestor-closed node set:
\begin{equation}
    \widetilde T(q)=\operatorname{PruneTree}\bigl(T(D),\widetilde R(q)\bigr).
\end{equation}

The resulting tree preserves parent--child relations among retained nodes
while removing unselected branches. The answer context $C(q,D)$ is formed
from their complete local source text and structural information. Semantic
summaries and atomic facts guide selection but do not substitute for the
source evidence. Benchmark-specific input formats are detailed in
Appendix~\ref{app:selection-implementation}.

Construction and selection require no method-specific parameter updates.
The hierarchy is reusable across queries, while each query induces its own
selected subtree. For a fixed query, the constructed context can be shared
across answer models. Algorithm~\ref{alg:selection} in
Appendix~\ref{app:selection-implementation} provides the complete selection
and pruning procedure.

\section{Experiments}
\label{sec:results}

\subsection{Experimental Setup}
\label{sec:setup}
\paragraph{Implementation.}
For \emph{source-grounded tree construction (\S\ref{sec:semantic-tree})}, we
use Qwen3.6-35B-A3B~\citep{qwen36modelcard2026} to generate the semantic hierarchy
and node descriptions, and a forward source matcher to align generated boundary
clues with original document spans. The resulting trees are cached for reuse.
For \emph{dual-granularity node selection (\S\ref{sec:dual-selection})}, the
semantic channel uses the same Qwen3.6-35B-A3B model to score nodes against the
query, while the lexical channel uses node-level BM25 scores obtained by taking
the maximum normalized score over each node's source-text subunits. For
\emph{structure-preserving pruning (\S\ref{sec:context-assembly})}, we take
the union of the two selections, add ancestor nodes, and prune unselected
branches while preserving parent--child links and complete local source text.
LongBench receives the resulting tree-structured JSON; LaRA formats the retained
nodes as numbered source chunks with ancestor-title provenance. We evaluate
these contexts with Qwen3.8-Max~\citep{qwen38max2026},
DeepSeek-V4-Pro~\citep{deepseekv4doc2026}, Kimi-K2.6~\citep{kimik26blog2026},
and GLM-5.3~\citep{glm53docs2026}, reusing each query's constructed context
across answer models without model-specific fine-tuning and holding it fixed
across three answer-generation runs. Construction and selection prompts,
selection settings, and input formats are documented in
Appendices~\ref{app:prompt-templates}, \ref{app:construction-selection},
and~\ref{app:selection-implementation}.

\begin{table*}[t]
\centering
\small
\caption{LongBench QA-F1 (three-run means). Overall averages four models;
parentheses show changes from Raw. Bold/underline mark the best/second-best means.}
\label{tab:longbench-main}
\fontsize{8.5}{10}\selectfont
\setlength{\tabcolsep}{1.8pt}
\renewcommand{\arraystretch}{1.1}
\begin{tabular*}{\textwidth}{@{\extracolsep{\fill}}llccccc@{}}
\toprule
Dataset & Method & Qwen3.8-Max & DeepSeek-V4-Pro & Kimi-K2.6 & GLM-5.3 & Overall \\
\midrule
\multirow{7}{*}{Qasper}
 & Raw & 48.89 & \underline{51.20} & \underline{50.35} & 46.61 & \underline{49.26} \\
 & LongLL~{\scriptsize\citep{longllmlingua2024}} & 29.75 & 26.55 & 24.58 & 27.72 & 27.15\,{\scriptsize (\drop{22.11})} \\
 & Provence~{\scriptsize\citep{provence2025}} & 47.50 & 46.18 & 48.17 & 46.84 & 47.17\,{\scriptsize (\drop{2.09})} \\
 & HiChunk~{\scriptsize\citep{hichunk2025}} & 49.70 & 49.88 & 49.24 & 46.85 & 48.92\,{\scriptsize (\drop{0.35})} \\
 & BRIEF-Pro~{\scriptsize\citep{briefpro2026}} & \underline{49.93} & 48.50 & 47.98 & \underline{47.06} & 48.37\,{\scriptsize (\drop{0.90})} \\
 & LongRefiner~{\scriptsize\citep{longrefiner2025}} & 31.52 & 31.00 & 29.91 & 29.59 & 30.50\,{\scriptsize (\drop{18.76})} \\
\noalign{\vskip 2pt}\cline{2-7}\noalign{\vskip 2pt}
 & \ours (ours) & \textbf{50.96} & \textbf{51.35} & \textbf{50.62} & \textbf{48.05} & \textbf{50.25}\,{\scriptsize (\gain{0.98})} \\
\midrule
\multirow{7}{*}{HotpotQA}
 & Raw & \underline{75.69} & \underline{72.20} & 59.47 & \textbf{70.68} & \underline{69.51} \\
 & LongLL~{\scriptsize\citep{longllmlingua2024}} & 67.69 & 55.48 & 49.11 & 53.92 & 56.55\,{\scriptsize (\drop{12.96})} \\
 & Provence~{\scriptsize\citep{provence2025}} & 71.44 & 66.21 & 58.05 & 68.35 & 66.01\,{\scriptsize (\drop{3.50})} \\
 & HiChunk~{\scriptsize\citep{hichunk2025}} & 70.53 & 69.57 & 58.16 & 68.21 & 66.62\,{\scriptsize (\drop{2.89})} \\
 & BRIEF-Pro~{\scriptsize\citep{briefpro2026}} & 73.80 & 69.61 & \underline{60.22} & 69.09 & 68.18\,{\scriptsize (\drop{1.33})} \\
 & LongRefiner~{\scriptsize\citep{longrefiner2025}} & 64.64 & 53.24 & 54.89 & 57.80 & 57.64\,{\scriptsize (\drop{11.87})} \\
\noalign{\vskip 2pt}\cline{2-7}\noalign{\vskip 2pt}
 & \ours (ours) & \textbf{77.39} & \textbf{73.64} & \textbf{63.50} & \underline{70.38} & \textbf{71.23}\,{\scriptsize (\gain{1.72})} \\
\midrule
\multirow{7}{*}{2WikiMQA}
 & Raw & \underline{84.19} & \underline{79.28} & 47.44 & \underline{71.62} & \underline{70.63} \\
 & LongLL~{\scriptsize\citep{longllmlingua2024}} & 75.87 & 59.16 & 35.61 & 55.53 & 56.54\,{\scriptsize (\drop{14.09})} \\
 & Provence~{\scriptsize\citep{provence2025}} & 78.87 & 55.32 & 46.15 & 66.97 & 61.82\,{\scriptsize (\drop{8.81})} \\
 & HiChunk~{\scriptsize\citep{hichunk2025}} & 80.65 & \textbf{79.54} & 47.75 & 70.79 & 69.69\,{\scriptsize (\drop{0.95})} \\
 & BRIEF-Pro~{\scriptsize\citep{briefpro2026}} & 81.70 & 70.22 & \underline{54.27} & 71.17 & 69.34\,{\scriptsize (\drop{1.29})} \\
 & LongRefiner~{\scriptsize\citep{longrefiner2025}} & 65.53 & 32.50 & 38.57 & 46.59 & 45.80\,{\scriptsize (\drop{24.84})} \\
\noalign{\vskip 2pt}\cline{2-7}\noalign{\vskip 2pt}
 & \ours (ours) & \textbf{84.85} & 79.00 & \textbf{54.44} & \textbf{74.08} & \textbf{73.09}\,{\scriptsize (\gain{2.46})} \\
\bottomrule
\end{tabular*}
\end{table*}


\paragraph{Benchmarks and metrics.}
We evaluate context construction across complementary QA settings and document
domains, with four frontier answer models testing the reuse of the same selected
contexts. Following prior compression research~\citep{longllmlingua2024},
LongBench~\citep{longbench2024} provides scientific-document QA through
Qasper~\citep{dasigi2021qasper} and multi-passage reasoning through
HotpotQA~\citep{yang2018hotpotqa} and 2WikiMultihopQA~\citep{ho2020twowiki}.
We follow its official answer templates and normalized token-overlap QA-F1
scoring. LaRA-32K~\citep{lara2025} complements these tasks with 887 questions
spanning 64 books, scientific papers, and financial reports, testing transfer
across document domains. We report accuracy using LaRA's official task-specific
grading prompts with a fixed gpt-oss-20b judge. All methods and answer models
share the same question set per dataset across three runs. Scores average the
runs; Overall equally averages the four model scores on this common set. The
LongBench summary weights its three tasks equally. Context reduction compares mean answer-context
character counts with Raw on aligned examples. Sample alignment is detailed in
Appendix~\ref{app:sample-alignment}.

\paragraph{Baselines.}
\textbf{Raw context} passes the uncompressed context and question to the answer
model. Baselines include \textbf{LongLLMLingua}~\citep{longllmlingua2024}
(\textbf{LongLL}) for query-aware token compression;
\textbf{Provence}~\citep{provence2025} for sentence pruning;
\textbf{HiChunk}~\citep{hichunk2025} for hierarchical chunking and retrieval;
\textbf{BRIEF-Pro}~\citep{briefpro2026} for generative compression; and adapted
\textbf{LongRefiner}~\citep{longrefiner2025} for query-aware document refinement.
We use reported construction settings where applicable. To compare performance
at similar mean context lengths, we also adjust baseline budgets or pruning
thresholds toward \ours's retained length, reporting achieved lengths and
scores. See Appendices~\ref{app:baseline-configurations}
and~\ref{app:length-adjusted-results} for details.

\subsection{Main Results}
\label{sec:longbench-results}
Tables~\ref{tab:longbench-main} and~\ref{tab:lara-main} summarize the LongBench
and LaRA comparisons under the evaluation protocol in \S\ref{sec:setup}. We
examine answering performance across tasks and answer models, followed by an
analysis of the evidence retained in the constructed contexts. Run-level
confidence intervals are reported in Appendix~\ref{app:main-result-details}.
Compression--performance analysis and comparisons at adjusted context lengths
are provided in Appendices~\ref{app:compression-tradeoff}
and~\ref{app:length-adjusted-results}.

\paragraph{Higher average QA-F1 with shorter answer contexts.}
\ours achieves the highest four-model average QA-F1 among the evaluated
methods on each of the three LongBench tasks. Across tasks, it improves over
Raw by \textbf{1.72 QA-F1 points while reducing answer-context length by
58.55\%}. The gains are larger on the two multi-hop tasks: \ours reaches
Overall scores of 71.23 on HotpotQA and 73.09 on 2WikiMQA, exceeding Raw by
1.72 and 2.46 points, respectively. On 2WikiMQA with Kimi-K2.6, the score rises
from 47.44 to 54.44---a \emph{7.00-point absolute improvement and a 14.76\%
relative gain}. The larger gains on tasks requiring information integration
across passages are consistent with the motivation for structure-conditioned
selection. These results directly address our research question: even when the
complete document is available to the answer model, selecting and organizing
its content can improve average answering performance rather than merely
reduce input length. Appendix~\ref{app:cost-benefit} quantifies when
answer-input savings offset amortized tree construction and query selection
under model-weighted input-token accounting.

\paragraph{Gains across answer models with shared contexts.}
\ours improves the LongBench average for \textbf{every evaluated answer model},
despite all four backbones already offering strong long-context capabilities.
The largest gain occurs on Kimi-K2.6, whose average rises from 52.42 to
56.19---an absolute improvement of 3.77 points and a relative gain of 7.19\%.
Qwen3.8-Max, DeepSeek-V4-Pro, and GLM-5.3 improve by 1.48, 0.44, and 1.20
points, respectively. This consistency across model families suggests that the
method is not tied to a particular backbone. These gains require
\emph{no answer-model-specific context construction or fine-tuning}: the same
prepared contexts are reused across models. They show that access to the full
document does not eliminate the value of upstream information organization
and selection.

\Needspace{20\baselineskip}
\begingroup
\setlength{\intextsep}{0pt}
\setlength{\columnsep}{0.03\textwidth}
\begin{wraptable}{r}{0.50\textwidth}
\centering
\setlength{\abovecaptionskip}{0pt}
\caption{LaRA accuracy (\%) averaged over three answer runs. Overall averages
the four models, with changes from Raw in parentheses.}
\label{tab:lara-main}
\fontsize{8}{9.5}\selectfont
\setlength{\tabcolsep}{0.5pt}
\renewcommand{\arraystretch}{1.1}
\begin{tabular*}{\linewidth}{@{\extracolsep{\fill}}lrrrrc@{}}
\toprule
Method & Qwen & DS & Kimi & GLM & Overall \\
\midrule
Raw & \underline{79.26} & \underline{76.52} & \underline{80.66} & \textbf{84.22} & \underline{80.17} \\
LongLL & 64.45 & 61.14 & 62.30 & 64.93 & 63.20\,(\drop{16.96}) \\
Provence & 71.85 & 70.77 & 71.93 & 76.96 & 72.88\,(\drop{7.29}) \\
HiChunk & 72.15 & 72.23 & 73.85 & 77.59 & 73.96\,(\drop{6.21}) \\
BRIEF-Pro & 72.30 & 71.14 & 70.73 & 75.31 & 72.37\,(\drop{7.79}) \\
LongRefiner & 43.07 & 38.79 & 38.68 & 43.04 & 40.89\,(\drop{39.27}) \\
\midrule
\ours & \textbf{79.37} & \textbf{78.14} & \textbf{80.89} & \underline{83.94} & \textbf{80.59}\,(\gain{0.42}) \\
\bottomrule
\end{tabular*}
\end{wraptable}
\paragraph{Cross-benchmark performance.}
LaRA extends the evaluation to heterogeneous document domains and question
types. Under the main-comparison settings in Table~\ref{tab:lara-main}, all
evaluated external baselines exhibit substantial declines in four-model average
accuracy relative to Raw, ranging from 6.21 to 39.27 points. In contrast,
\ours achieves the \textbf{highest four-model average accuracy with 43.87\%
less answer context}, increasing accuracy from 80.17 to 80.59. DeepSeek-V4-Pro
gains 1.62 accuracy points over Raw. This contrast shows that substantial
context reduction need not come at the expense of average answer quality in
this setting. Together with the LongBench results, it supports the applicability
of source-grounded context construction across different long-document
evaluation settings.
\par
\endgroup

\paragraph{Evidence completeness and concentration.}
\label{sec:ecp-analysis}
\begingroup
\setlength{\intextsep}{2pt}
\setlength{\columnsep}{0.025\textwidth}
\begin{wraptable}{r}{0.54\textwidth}
\centering
\setlength{\abovecaptionskip}{0pt}
\setlength{\belowcaptionskip}{5pt}
\footnotesize
\caption{Evidence coverage, purity, and ECP (\%). Overall QA-F1 is reused
from Table~\ref{tab:longbench-main}.}
\label{tab:evidence-components-main}
\label{tab:evidence-ecp}
\fontsize{6.6}{7.2}\selectfont
\setlength{\tabcolsep}{1.1pt}
\renewcommand{\arraystretch}{0.94}
\begin{tabular*}{\linewidth}{@{\extracolsep{\fill}}llrrrr@{}}
\toprule
Dataset & Method & All-hit & Purity & ECP & Overall \\
\midrule
\multirow{6}{*}{Qasp.}
 & LongLL & 0.57 & 2.18 & 1.38 & 27.15 \\
 & Prov. & 16.57 & \underline{17.82} & 17.20 & 47.17 \\
 & HiChunk & \underline{96.00} & 5.80 & \underline{50.90} & \underline{48.92} \\
 & BRIEF & 73.71 & \textbf{25.41} & 49.56 & 48.37 \\
 & LongRef. & 20.57 & 5.19 & 12.88 & 30.50 \\
 & \ours & \textbf{96.57} & 9.33 & \textbf{52.95} & \textbf{50.25} \\
\midrule
\multirow{6}{*}{Hotpot}
 & LongLL & 0.00 & 28.95 & 14.48 & 56.55 \\
 & Prov. & 0.00 & 47.67 & 23.84 & 66.01 \\
 & HiChunk & \underline{12.23} & 52.42 & 32.33 & 66.62 \\
 & BRIEF & 0.00 & \textbf{92.61} & \textbf{46.31} & \underline{68.18} \\
 & LongRef. & 0.00 & 49.00 & 24.50 & 57.64 \\
 & \ours & \textbf{12.77} & \underline{71.44} & \underline{42.11} & \textbf{71.23} \\
\midrule
\multirow{6}{*}{2Wiki}
 & LongLL & 2.55 & 35.53 & 19.04 & 56.54 \\
 & Prov. & 0.00 & 52.81 & 26.41 & 61.82 \\
 & HiChunk & \textbf{55.10} & 43.28 & 49.19 & \underline{69.69} \\
 & BRIEF & 8.16 & \textbf{97.93} & \underline{53.05} & 69.34 \\
 & LongRef. & 0.51 & 34.81 & 17.66 & 45.80 \\
 & \ours & \underline{41.33} & \underline{82.69} & \textbf{62.01} & \textbf{73.09} \\
\bottomrule
\end{tabular*}
\end{wraptable}
Table~\ref{tab:evidence-components-main} relates annotated evidence preservation
to answer quality; the \emph{Overall} column directly reuses the four-model
QA-F1 values from Table~\ref{tab:longbench-main}. All-hit measures complete
coverage under the evidence-matching rule, Evidence Purity measures the
concentration of selected content on annotated source evidence, and ECP is their
equally weighted mean. The evidence metrics use annotated subsets, with
definitions and matching procedures provided in
Appendices~\ref{app:gold-alignment}--\ref{app:ecp-uncertainty}.
\ours achieves the highest Overall on all three datasets, together with the
\textbf{highest ECP on Qasper (52.95) and 2WikiMQA (62.01)} and the
second-highest on HotpotQA (42.11). The component scores clarify these ECP
results. On Qasper, \ours preserves nearly the same complete-evidence coverage
as HiChunk (96.57\% versus 96.00\%) while achieving higher purity (9.33\%
versus 5.80\%). On HotpotQA, it exceeds HiChunk on both measures. BRIEF-Pro's
higher purity gives it the highest ECP on this task, whereas \ours obtains
higher Overall QA-F1 (71.23 versus 68.18). On 2WikiMQA, \ours combines the
second-highest All-hit and purity to attain the highest ECP. These comparisons
characterize a combination of support retention and concentration, rather than
dominance in either component alone. This pattern is consistent with the design
of selecting structurally relevant source nodes and supplementing them with
precise lexical matches while preserving the local context of retained evidence.
\par
\endgroup

\subsection{Ablation Studies}
\label{sec:ablation}
We examine how semantic and lexical selection complement each other, what
semantic units and hierarchy contribute, and how tree construction affects
answering. Tables~\ref{tab:ablation-selection}--\ref{tab:ablation-construction}
compare variants on HotpotQA and 2WikiMQA, both requiring cross-passage evidence
integration. Qwen3.8-Max answers questions; Qwen3.6-35B-A3B constructs and
scores nodes. Scores average three runs. Avg.\ is the unweighted two-task mean,
with changes from Raw in parentheses. BM approximately matches mean retained
characters, not per-question token counts. Best means are bold and runners-up
underlined. See Appendix~\ref{app:ablation-variants} for implementation details.

\Needspace{15\baselineskip}
\begingroup
\setlength{\intextsep}{0pt}
\setlength{\columnsep}{0.03\textwidth}
\setlength{\belowcaptionskip}{5pt}
\begin{wraptable}{r}{0.50\textwidth}
\centering
\setlength{\abovecaptionskip}{0pt}
\caption{Effects of semantic and lexical selection.}
\label{tab:ablation-summary}
\label{tab:ablation-selection}
\fontsize{8}{9.5}\selectfont
\setlength{\tabcolsep}{0.8pt}
\renewcommand{\arraystretch}{1.1}
\begin{tabular*}{\linewidth}{@{\extracolsep{\fill}}lrrr@{}}
\toprule
Variant & HotpotQA & 2WikiMQA & Avg. \\
\midrule
Raw & 75.69 & \underline{84.19} & 79.94 \\
Semantic-only & \underline{76.59} & 83.97 & \underline{80.28} (\gain{0.34}) \\
Lexical-only (BM) & 73.54 & 76.38 & 74.96 (\drop{4.98}) \\
Semantic-only (BM) & 74.93 & 81.87 & 78.40 (\drop{1.54}) \\
\midrule
\ours & \textbf{77.39} & \textbf{84.85} & \textbf{81.12} (\gain{1.18}) \\
\bottomrule
\end{tabular*}
\end{wraptable}

\paragraph{Semantic and lexical selection.}
We test \emph{whether the two selection channels contribute beyond increasing
context length} by comparing three variants in
Table~\ref{tab:ablation-selection}. \emph{Semantic-only} retains semantic
selections and their ancestor context, removing lexical supplementation.
\emph{Lexical-only (BM)} uses lexical node selection with ancestor closure
under an approximately matched character budget. \emph{Semantic-only (BM)}
instead supplements semantic selections with random nodes using a single fixed
seed, approaching the full method's context length without targeted lexical
matching. \ours reaches \textbf{81.12 average QA-F1}, exceeding Semantic-only
by 0.84 points, Lexical-only (BM) by 6.16 points, and Semantic-only (BM) by
2.72 points. The smaller difference from Semantic-only indicates that semantic
selection accounts for most of the observed performance, with lexical matching
providing an incremental contribution. On HotpotQA and 2WikiMQA, the full
method exceeds Semantic-only by 0.81 and 0.88 points, whereas random
supplementation scores 1.66 and 2.10 points lower than Semantic-only. These
contrasts favor \textbf{targeted supplementation over the tested random-expansion
configuration}, rather than attributing the improvement simply to additional
context.

To examine whether these performance differences are accompanied by differences
in evidence retention, we match each variant's saved context against the mapped
gold annotations used in our source-evidence analysis
(Appendix~\ref{app:gold-alignment}). Span Recall measures length-weighted
coverage of annotated source evidence. Table~\ref{tab:conditional-evidence}
in Appendix~\ref{app:conditional-ablation} reports the conditional results.
Its pooled column shows that, on the annotated subsets across all three tasks,
including Qasper, \ours has 21.50 percentage points higher Span Recall and
4.45 points higher QA-F1 than Lexical-only (BM), pooling over questions.
Relative to the original Semantic-only configuration, the corresponding
differences are 1.17 percentage points and 0.36 QA-F1. These contrasts are
consistent with structural--semantic selection providing most evidence coverage
and lexical matching serving as a smaller, targeted supplement. The HotpotQA
and 2WikiMQA columns of Table~\ref{tab:conditional-evidence} also show that
the full method obtains 2.42 and 2.76 points higher QA-F1 than Semantic-only
(BM) on their annotated subsets, despite small differences in aggregate Span
Recall. This pattern emphasizes \emph{which source details are retained},
rather than treating additional context as interchangeable.

\par
\endgroup

\begingroup
\setlength{\intextsep}{0pt}
\setlength{\columnsep}{0.03\textwidth}
\begin{wraptable}{r}{0.50\textwidth}
\centering
\setlength{\abovecaptionskip}{0pt}
\caption{Effects of semantic units and hierarchy.}
\label{tab:ablation-hierarchy}
\fontsize{8}{9.5}\selectfont
\setlength{\tabcolsep}{0.8pt}
\renewcommand{\arraystretch}{1.1}
\begin{tabular*}{\linewidth}{@{\extracolsep{\fill}}lrrr@{}}
\toprule
Variant & HotpotQA & 2WikiMQA & Avg. \\
\midrule
Raw & \underline{75.69} & \underline{84.19} & \underline{79.94} \\
Fixed 200 & 72.77 & 79.47 & 76.12 (\drop{3.82}) \\
Chonkie & 74.33 & 78.77 & 76.55 (\drop{3.39}) \\
\ours (flat) & 74.07 & 80.93 & 77.50 (\drop{2.44}) \\
\midrule
\ours & \textbf{77.39} & \textbf{84.85} & \textbf{81.12} (\gain{1.18}) \\
\bottomrule
\end{tabular*}
\end{wraptable}

\paragraph{Semantic units and hierarchy.}
We examine \emph{what semantic units and hierarchical organization contribute}
through two types of comparisons in Table~\ref{tab:ablation-hierarchy}.
First, we compare \ours with \emph{\ours{} (flat)}, which retains the same
source-node inventory but removes the hierarchy. This comparison evaluates the joint use of
hierarchy-aware scoring, ancestor completion, and structured serialization.
The selected content and retained length can also change, so it is not a
same-context test of structure alone. We additionally compare \emph{Fixed 200}
and \emph{Chonkie}, which replace the semantic units with fixed 200-token
chunks and Chonkie semantic chunks, respectively. These representation-level
comparisons change unit boundaries and metadata, complementing the
same-inventory hierarchy comparison. The hierarchy-enabled pipeline achieves
\textbf{81.12 average QA-F1 versus 77.50 for the flat pipeline}, a 3.62-point
difference across the two multi-hop tasks. The differences are 3.32 points on
HotpotQA and 3.92 points on 2WikiMQA. Fixed 200 and Chonkie yield average
scores of 76.12 and 76.55, falling below the full method by 5.00 and 4.57
points, respectively. The same-inventory comparison is consistent with
hierarchical processing being useful for selecting support distributed across
passages, while the alternative-unit comparisons support the use of the
source-grounded semantic representation within this selection framework.

\par
\endgroup

\Needspace{14\baselineskip}
\begingroup
\setlength{\intextsep}{0pt}
\setlength{\columnsep}{0.03\textwidth}
\begin{wraptable}{r}{0.50\textwidth}
\centering
\setlength{\abovecaptionskip}{0pt}
\caption{Effects of tree-construction strategy.}
\label{tab:ablation-construction}
\fontsize{8}{9.5}\selectfont
\setlength{\tabcolsep}{0.8pt}
\renewcommand{\arraystretch}{1.1}
\begin{tabular*}{\linewidth}{@{\extracolsep{\fill}}lrrr@{}}
\toprule
Variant & HotpotQA & 2WikiMQA & Avg. \\
\midrule
Raw & \underline{75.69} & \underline{84.19} & \underline{79.94} \\
Multi-stage & 75.47 & 82.33 & 78.90 (\drop{1.04}) \\
HiChunk adapter & 74.15 & 82.54 & 78.34 (\drop{1.60}) \\
\midrule
\ours & \textbf{77.39} & \textbf{84.85} & \textbf{81.12} (\gain{1.18}) \\
\bottomrule
\end{tabular*}
\end{wraptable}

\paragraph{Tree-construction strategies.}
We ask \emph{how joint tree construction compares with staged and adapted
representations}. Table~\ref{tab:ablation-construction} compares these strategies
under the same downstream selection framework. The \emph{Multi-stage} configuration
uses the same Qwen3.6-35B-A3B model but separates construction into three
stages: extracting and extending a document outline from grouped source chunks;
validating section boundaries and anchoring them to the original text; and
enriching the resulting nodes with summaries and facts. The \emph{HiChunk
adapter} maps HiChunk-derived units into the node representation used by our
selector, rather than evaluating the complete HiChunk retrieval system.
Implementation and configuration details are provided in
Appendix~\ref{app:construction-variants}. Within the tested selector,
Qwen3.8-Max answer model, and two multi-hop tasks, the single-pass constructor
achieves \textbf{81.12 average QA-F1}, compared with 78.90 for Multi-stage
and 78.34 for the HiChunk adapter. Its advantage over Multi-stage is 2.22
points on average, with differences of 1.92 points on HotpotQA and 2.52 points
on 2WikiMQA. These results demonstrate that the tested single-generation
constructor can provide an effective representation for this selection pipeline.
Together with the hierarchy comparison in Table~\ref{tab:ablation-hierarchy},
indicating that using document structure effectively does not require a more elaborate construction procedure.
\par
\endgroup

\section{Conclusion}
We revisited retrieval in the long-context era as a problem of selecting and
organizing evidence, rather than merely fitting documents into a limited window.
\ours operationalizes this view through a training-free, reusable semantic
hierarchy: abstraction guides selection, while structure-preserving pruning
retains original text for answering. Across four frontier answer models, \ours
improves average QA-F1 by 1.72 points over full-context prompting on three
LongBench tasks with 58.55\% less answer context, and achieves the highest
four-model average accuracy on LaRA with 43.87\% less context. Under the main
evaluation settings, all compared retrieval and compression baselines fall below
Raw in four-model average performance on each LongBench task and LaRA, whereas
\ours exceeds it. This contrast validates the value of adaptive context
construction while showing that input reduction alone does not ensure better
answers. The central design principle is to use abstraction to decide what to
retain without substituting it for the source evidence. Long-context capacity
and retrieval thus address complementary decisions: how much information can be
made available, and which source content and relations should be presented for
a particular question. Our findings establish the latter as a consequential
design choice even when the former no longer prevents full-document input.

\paragraph{Limitation and Future Work.}
Our evaluation centers on long-document question answering. Extending \ours to
multimodal documents would require linking textual evidence with tables and
figures. Although retained text is source-grounded, selection still depends on
generated structure and descriptions. Controlled perturbation studies could
quantify this sensitivity and guide more robust tree construction and selection.
Future work could also explore other evidence-intensive tasks and incremental
tree updates as documents evolve, extending the reusable hierarchy beyond the
settings evaluated here.

\section*{AI Use Statement}

In this work, we used generative AI tools to assist with writing,
specifically for language polishing, grammar correction, and improving the
clarity of exposition in text that was first drafted by the authors. We also
used generative AI tools as coding assistants for routine, non-novel
implementation tasks (e.g., boilerplate code, plotting scripts, and refactoring).
We have not used generative AI tools to generate research ideas, formulate the
core hypotheses, design the method, produce or analyze experimental results, or
draw scientific conclusions; these were carried out entirely by the authors.

\section*{Ethics Statement}
\textsc{TreeFuse} uses public benchmarks and involves no human-subject studies or collection of new personal data. Dataset and model use follows their respective terms. Source-preserving selection does not guarantee factual correctness or privacy protection; sensitive applications require appropriate safeguards.

\section*{Reproducibility Statement}
Our code is available through the code repository linked in the paper. The appendices document prompts, selection rules, experimental settings, and evaluation procedures. Fixed contexts are reused across answer models and repeated runs; API updates and stochastic generation may affect exact replication.

\bibliography{references}
\bibliographystyle{src/iclr2027_conference}

\clearpage
\appendix
\section*{Appendix Contents}
\label{app:contents}
\begingroup
\small
\setlength{\parskip}{1pt}
\medskip
\noindent \textbf{\hyperref[app:method-details]{\ref*{app:method-details}~Method Details and Relation to Prior Work}}\nobreak\hfill \hyperref[app:method-details]{\pageref*{app:method-details}}\par
\noindent \hspace*{1em}\hyperref[app:related-comparison]{\ref*{app:related-comparison}~Comparison with Related Methods}\nobreak\hfill \hyperref[app:related-comparison]{\pageref*{app:related-comparison}}\par
\noindent \hspace*{1em}\hyperref[app:prompt-templates]{\ref*{app:prompt-templates}~Prompt Templates and Output Schemas}\nobreak\hfill \hyperref[app:prompt-templates]{\pageref*{app:prompt-templates}}\par
\noindent \hspace*{1em}\hyperref[app:construction-selection]{\ref*{app:construction-selection}~Tree Construction, Scoring, and Selection}\nobreak\hfill \hyperref[app:construction-selection]{\pageref*{app:construction-selection}}\par
\noindent \hspace*{1em}\hyperref[app:selection-implementation]{\ref*{app:selection-implementation}~Structure-Preserving Serialization}\nobreak\hfill \hyperref[app:selection-implementation]{\pageref*{app:selection-implementation}}\par
\medskip
\noindent \textbf{\hyperref[app:protocol-audit]{\ref*{app:protocol-audit}~Experimental Configurations and Evaluation}}\nobreak\hfill \hyperref[app:protocol-audit]{\pageref*{app:protocol-audit}}\par
\noindent \hspace*{1em}\hyperref[app:sample-alignment]{\ref*{app:sample-alignment}~Benchmarks and Sample Alignment}\nobreak\hfill \hyperref[app:sample-alignment]{\pageref*{app:sample-alignment}}\par
\noindent \hspace*{1em}\hyperref[app:answer-configurations]{\ref*{app:answer-configurations}~Answer Models and Judge Configurations}\nobreak\hfill \hyperref[app:answer-configurations]{\pageref*{app:answer-configurations}}\par
\noindent \hspace*{1em}\hyperref[app:baseline-configurations]{\ref*{app:baseline-configurations}~Baseline Implementations and Context Budgets}\nobreak\hfill \hyperref[app:baseline-configurations]{\pageref*{app:baseline-configurations}}\par
\medskip
\noindent \textbf{\hyperref[app:supplementary-results]{\ref*{app:supplementary-results}~Supplementary Quantitative Results}}\nobreak\hfill \hyperref[app:supplementary-results]{\pageref*{app:supplementary-results}}\par
\noindent \hspace*{1em}\hyperref[app:main-result-details]{\ref*{app:main-result-details}~Main-Comparison Details and Run-Level Intervals}\nobreak\hfill \hyperref[app:main-result-details]{\pageref*{app:main-result-details}}\par
\noindent \hspace*{1em}\hyperref[app:compression-tradeoff]{\ref*{app:compression-tradeoff}~Compression--Performance Trade-off}\nobreak\hfill \hyperref[app:compression-tradeoff]{\pageref*{app:compression-tradeoff}}\par
\noindent \hspace*{1em}\hyperref[app:length-adjusted-results]{\ref*{app:length-adjusted-results}~Comparisons with Adjusted Context Lengths}\nobreak\hfill \hyperref[app:length-adjusted-results]{\pageref*{app:length-adjusted-results}}\par
\noindent \hspace*{1em}\hyperref[app:cost-benefit]{\ref*{app:cost-benefit}~Context Savings and Construction Amortization}\nobreak\hfill \hyperref[app:cost-benefit]{\pageref*{app:cost-benefit}}\par
\medskip
\noindent \textbf{\hyperref[app:evidence-metrics]{\ref*{app:evidence-metrics}~Source-Evidence Analysis}}\nobreak\hfill \hyperref[app:evidence-metrics]{\pageref*{app:evidence-metrics}}\par
\noindent \hspace*{1em}\hyperref[app:gold-alignment]{\ref*{app:gold-alignment}~Evidence Annotations and Source Alignment}\nobreak\hfill \hyperref[app:gold-alignment]{\pageref*{app:gold-alignment}}\par
\noindent \hspace*{1em}\hyperref[app:evidence-definitions]{\ref*{app:evidence-definitions}~Evidence Metric Definitions}\nobreak\hfill \hyperref[app:evidence-definitions]{\pageref*{app:evidence-definitions}}\par
\noindent \hspace*{1em}\hyperref[app:ecp-uncertainty]{\ref*{app:ecp-uncertainty}~Equal-Weight ECP Rationale}\nobreak\hfill \hyperref[app:ecp-uncertainty]{\pageref*{app:ecp-uncertainty}}\par
\medskip
\noindent \textbf{\hyperref[app:ablations]{\ref*{app:ablations}~Ablation Details and Case Studies}}\nobreak\hfill \hyperref[app:ablations]{\pageref*{app:ablations}}\par
\noindent \hspace*{1em}\hyperref[app:ablation-variants]{\ref*{app:ablation-variants}~Selection, Representation, and Construction Variants}\nobreak\hfill \hyperref[app:ablation-variants]{\pageref*{app:ablation-variants}}\par
\noindent \hspace*{1em}\hyperref[app:construction-variants]{\ref*{app:construction-variants}~Tree-Construction Configurations}\nobreak\hfill \hyperref[app:construction-variants]{\pageref*{app:construction-variants}}\par
\noindent \hspace*{1em}\hyperref[app:conditional-ablation]{\ref*{app:conditional-ablation}~Conditional Evidence and Answer Analysis}\nobreak\hfill \hyperref[app:conditional-ablation]{\pageref*{app:conditional-ablation}}\par
\noindent \hspace*{1em}\hyperref[app:selection-cases]{\ref*{app:selection-cases}~Semantic Support and Lexical Supplementation Cases}\nobreak\hfill \hyperref[app:selection-cases]{\pageref*{app:selection-cases}}\par
\noindent \hspace*{1em}\hyperref[app:boundary-adequacy-case]{\ref*{app:boundary-adequacy-case}~Semantic Adequacy versus Boundary Exactness}\nobreak\hfill \hyperref[app:boundary-adequacy-case]{\pageref*{app:boundary-adequacy-case}}\par
\endgroup
\clearpage
\section{Method Details and Relation to Prior Work}
\label{app:method-details}
\subsection{Comparison with Related Methods}
\label{app:related-comparison}
Table~\ref{tab:related-comparison} compares \ours with evaluated baselines and
selected related methods in document representation, query-time selection, and
the evidence supplied for answering.

\begin{table*}[t]
\centering
\small
\caption{Comparison with evaluated baselines and selected related methods.
`Reuse' denotes a document representation shared across queries. `Train'
denotes method-specific parameter updates.}
\label{tab:related-comparison}
\renewcommand{\arraystretch}{1.12}
\setlength{\tabcolsep}{2.5pt}
\begin{tabularx}{\textwidth}{@{}l>{\raggedright\arraybackslash}p{1.05in}>{\raggedright\arraybackslash}X>{\raggedright\arraybackslash}p{1.00in}c@{}}
\toprule
Method & Indexed representation & Query-time mechanism & Reader input & Reuse / Train \\
\midrule
LongLLMLingua & Flat prompt & Token pruning & Retained tokens & No / No \\
Provence & Flat passages & Sentence pruning & Source sentences & No / Yes \\
RAPTOR & Summary tree & Dense node retrieval & Summaries / nodes & Yes / No \\
LongRefiner & Document hierarchy & Learned adaptive refinement & Refined spans & Yes / Yes \\
HiChunk & Learned hierarchy & Retrieval + Auto-Merge & Merged chunks & Yes / Yes \\
ACC-RAG & Multi-granular embeddings & Adaptive compression rate & Compressed features & Yes / Yes \\
SARA & Text + semantic vectors & Fixed-budget reranking & Snippets + vectors & Yes / Yes \\
BRIEF-Pro & Flat passages & Generative compression & Synthesized brief & No / Yes \\
\midrule
\ours & \textbf{Semantic hierarchy linked to local source spans} & \textbf{Structure-conditioned scoring and lexical attribution over shared nodes} & \textbf{Pruned hierarchy with complete local source text} & \textbf{Yes / No} \\
\bottomrule
\end{tabularx}
\end{table*}

\begin{figure}[!t]
\centering
\includegraphics[width=\textwidth]{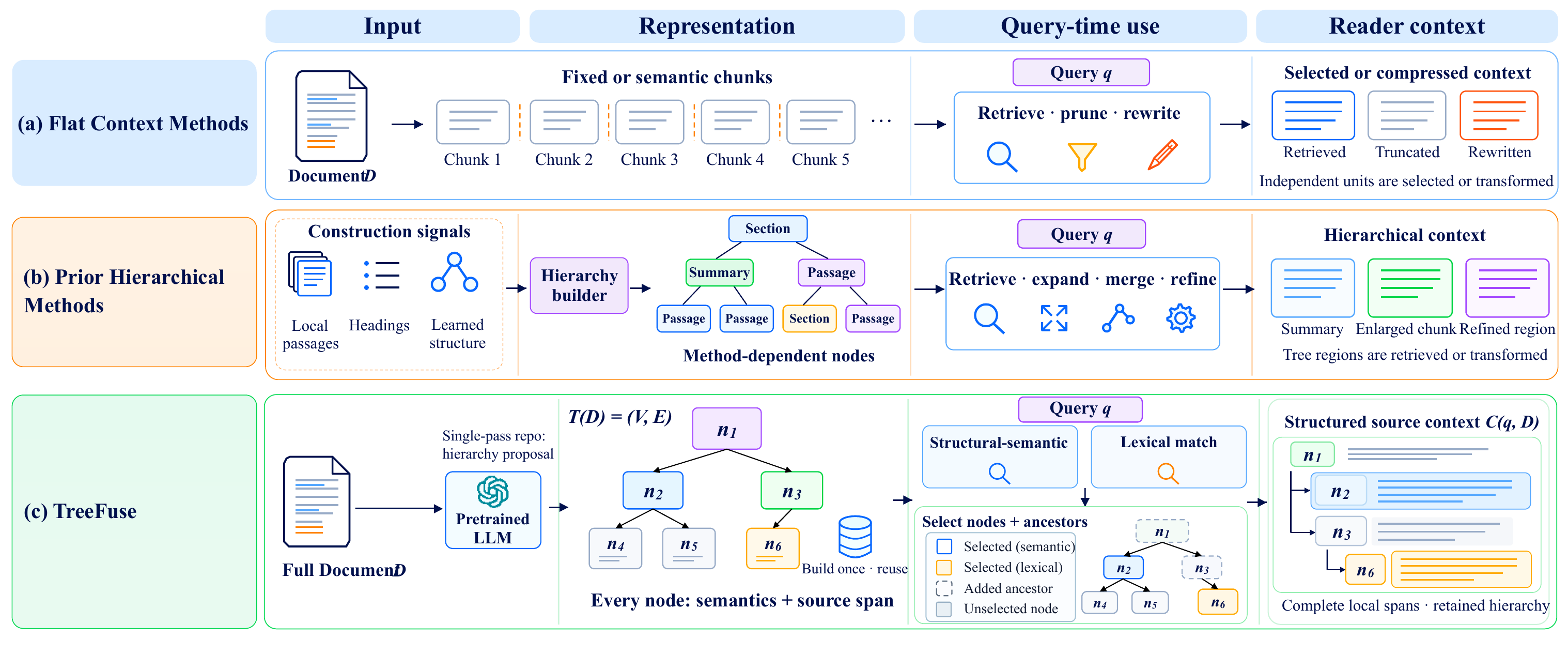}
\caption{Three paradigms for constructing an answer context. Flat methods organize text as independent units and retrieve, prune, or rewrite them; prior hierarchical methods build method-dependent trees and retrieve, expand, merge, or refine tree regions; TreeFuse constructs a reusable source-grounded semantic hierarchy from the full document, prunes it to an ancestor-closed retained subtree for each query, and serializes the complete local source spans while preserving the original parent–child relations and sibling order.
}
\label{fig:three-paradigms}
\end{figure}

\subsection{Prompt Templates and Output Schemas}
\label{app:prompt-templates}

The following condensed prompt descriptions summarize the instructions and
active interfaces. They are presentation summaries, not verbatim replacements
for the implementation templates; see the code repository linked in the paper.

\begin{table}[htbp]
\centering
\small
\caption{Principal interfaces used for construction, selection, and answering.
Only fields relevant to the all-node selection path are shown.}
\label{tab:prompt-interfaces}
\begin{tabularx}{\textwidth}{@{}p{0.24\textwidth}>{\raggedright\arraybackslash}X@{}}
\toprule
Interface & Fields \\
\midrule
Construction & \texttt{sections}: records with \texttt{structure},
\texttt{title}, \texttt{boundary\_type}, \texttt{heading\_verbatim},
\texttt{text\_before\_cut}, \texttt{summary}, and \texttt{facts}. \\
Scoring input & Query, document directory, and node records containing
\texttt{node\_id}, \texttt{title}, and semantic \texttt{material}. \\
Scoring output & \texttt{scores}: records with \texttt{node\_id}, integer
\texttt{relevance} in $[0,10]$, and \texttt{reason}. \\
Selection cutoff & Integer \texttt{keep\_threshold} in $[1,10]$. \\
Answer context & \texttt{doc\_id} and \texttt{tree}, with \texttt{level},
\texttt{title}, \texttt{source\_text}, and optional \texttt{children} per node. \\
\bottomrule
\end{tabularx}
\end{table}

\paragraph{Tree construction.}
\begin{Verbatim}[fontsize=\footnotesize,breaklines=true,frame=single,framesep=2mm]
Organize the full document into a hierarchical section tree in one pass.
Use headings and coherent topic changes as boundaries. Keep passages
separate and sections in document order. Give each section a unique
hierarchical path with its parent present. Copy boundary clues verbatim.
Summarize only the section's own text and request at most eight atomic,
source-supported facts, preserving exact entities, dates, and numbers.
Return a JSON object containing the sections and required fields.
Document: {document_text}
\end{Verbatim}

\paragraph{Node relevance.}
\begin{Verbatim}[fontsize=\footnotesize,breaklines=true,frame=single,framesep=2mm]
Given the query, document directory, and node metadata, score every node
from 0 to 10 for its contribution to answering. Use titles, summaries,
atomic facts requested during construction, and available parent context.
Judge direct evidence, supporting relations, comparisons, temporal facts,
and cross-node links rather than word overlap alone. Give high scores to
necessary support even when its wording differs from the question.
Return {"scores":[{"node_id":"...","relevance":0,"reason":"..."}]}.
Query: {query}; Directory: {directory}; Nodes: {nodes}
\end{Verbatim}

\paragraph{Selection cutoff.}
\begin{Verbatim}[fontsize=\footnotesize,breaklines=true,frame=single,framesep=2mm]
Choose a keep threshold from 1 to 10 using the query, scored outline,
score histogram, and retention profile. Preserve plausible supporting
nodes while excluding unrelated branches. Return the numeric cutoff.
Query: {query}; Outline: {tree_outline}; Histogram: {histogram};
Retention profile: {keep_curve}; Node count: {total}
\end{Verbatim}

\subsection{Tree Construction, Scoring, and Selection}
\label{app:construction-selection}
\paragraph{Tree construction and source alignment.}
\label{app:tree-construction}
Qwen3.6-35B-A3B proposes section paths, titles, summaries, atomic facts, and
verbatim boundary clues. A forward source matcher resolves boundaries
from headings, boundary text, or titles. Consecutive offsets delimit each node's
local text, with the first span starting at zero and the last ending at the
document boundary. Parent paths define the hierarchy without duplicating child
text inside parent nodes.

\paragraph{Scoring inputs.}
The scorer uses titles, summaries, atomic facts, and available parent context.
The construction request limits these facts to eight per section. When both
summary and facts are absent, the scorer uses a short source-text prefix.
Breadth-first batches include a document directory and score all nodes before
pruning. Table~\ref{tab:method-settings} summarizes the implementation settings.

\begin{table}[htbp]
\centering
\small
\caption{Method settings. Entries marked as defaults refer to the implementation.}
\label{tab:method-settings}
\begin{tabularx}{\textwidth}{@{}p{0.23\textwidth}>{\raggedright\arraybackslash}X p{0.21\textwidth}@{}}
\toprule
Setting & Value or behavior & Provenance \\
\midrule
Constructor / scorer & Qwen3.6-35B-A3B & Reported setting \\
Construction facts & Typically 3--6, at most 8 requested & Construction prompt \\
Scoring material & Section summary, atomic facts, and parent context & Node metadata \\
Missing metadata & First 300 source characters when both summary and facts are absent & Scorer implementation \\
Batch size & 40 nodes and 48,000 serialized-entry characters & Scorer defaults \\
Cutoff suggestions & 3 attempts, minimum valid keep threshold & Selector default \\
Coverage floors & 2\% / 3 nodes, or 5\% / 5 nodes for detected location queries & Selector defaults \\
Lexical cutoff & P90 for all datasets & Selector defaults \\
Node output & Complete local source text with retained ancestor structure & Serialization interface \\
\bottomrule
\end{tabularx}
\end{table}

\paragraph{Threshold selection and safeguards.}
\label{app:selection-thresholds}
\paragraph{Semantic cutoff.}
Node ratings lie in $[0,10]$. The selector requests three cutoff suggestions and
uses the minimum valid value in $[1,10]$. It lowers this cutoff as needed to
approach the coverage floor, including ancestor context. Default floors are
2\% of source characters and three nodes, increasing to 5\% and five nodes for
detected location questions. A fallback handles missing cutoff responses, and
an empty selection is replaced by the highest-scoring root and its subtree.

\paragraph{Lexical cutoff.}
BM25 is computed over a shared pool of source-text subunits. Short fragments
below 40 characters are merged into their preceding unit when one exists.
For nonconstant scores, normalization is
$z_u=(b_u-b_{\min})/(b_{\max}-b_{\min})$ over the complete document pool.
Node scores are the maxima of their subunit scores, with zero assigned to a
node without subunits. For every dataset, the lexical cutoff is the 90th
percentile (P90) of the node-score distribution. The percentile uses linear
interpolation and includes ties, so it does not select exactly 10\% of nodes.
Ancestor closure can further enlarge the set.

\subsection{Structure-Preserving Serialization}
\label{app:selection-implementation}
\textsc{PruneTree} retains selected nodes and their ancestors without changing
parent--child links or sibling order. LongBench serializes this tree as nested
records containing titles, hierarchy labels, and complete local source text.
LaRA uses numbered source chunks with ancestor-title provenance. The same
selected content is reused across answer models. The benchmark-specific output
format is denoted by $\rho$ in Algorithm~\ref{alg:selection}.

\begin{algorithm}[t]
\caption{\ours query-adaptive tree selection and serialization}
\label{alg:selection}
\begin{algorithmic}[1]
\STATE \textbf{Input:} cached hierarchy $T(D)=(V,E)$, query $q$
\STATE \textbf{Settings:} cutoff policies $\Pi_{\mathrm{sem}},\Pi_{\mathrm{lex}}$, serialization rule $\rho$
\STATE $r^{\mathrm{sem}}\gets\operatorname{StructSemScore}(T(D),q)$
\STATE $\tau\gets\Pi_{\mathrm{sem}}(q,T(D),r^{\mathrm{sem}})$
\STATE $R_{\mathrm{sem}}^{0}\gets\{n_i\in V:r_i^{\mathrm{sem}}\geq\tau\}$
\STATE $R_{\mathrm{sem}}\gets\operatorname{EnsureNonempty}(R_{\mathrm{sem}}^{0},T(D),r^{\mathrm{sem}})$
\STATE $\mathcal U(n_i)\gets\operatorname{Split}(x_i)$ for every $n_i\in V$
\STATE $\mathcal U\gets\bigcup_{n_i\in V}\mathcal U(n_i)$
\STATE $z_u\gets\operatorname{NormBM25}(q,u,\mathcal U)$ for every $u\in\mathcal U$
\STATE $b_i\gets\max\bigl(\{z_u:u\in\mathcal U(n_i)\}\cup\{0\}\bigr)$ for every $n_i\in V$
\STATE $\eta\gets\Pi_{\mathrm{lex}}(q,D,\{b_i\}_{n_i\in V})$
\STATE $R_{\mathrm{lex}}\gets\{n_i\in V:b_i\geq\eta\}$
\STATE $S\gets R_{\mathrm{sem}}\cup R_{\mathrm{lex}}$
\STATE $\widetilde{R}\gets S\cup\bigcup_{n_i\in S}\operatorname{Anc}_{T}(n_i)$
\STATE $\widetilde T\gets\operatorname{PruneTree}(T(D),\widetilde R)$
\STATE \textbf{Return} $C(q,D)\gets\operatorname{SerializeTree}(\widetilde T,\rho)$
\end{algorithmic}
\end{algorithm}

\paragraph{Serialized input example.}
This excerpt from a saved Qasper context preserves its actual field names,
titles, and source prefixes. Ellipses omit text for display only. Gold answers
and relevance scores are not part of the answer-model input.
\begin{Verbatim}[fontsize=\footnotesize,breaklines=true,breakanywhere=true,frame=single,framesep=2mm]
{"doc_id":"qasper-1-1704f1d50b2554ae","tree":[
  {"level":"level2","title":"Pooling Strategies",
   "source_text":"POOLING STRATEGIES\nIn any language identification model, we want to obtain utterance level representation ...",
   "children":[
     {"level":"level2.1","title":"NetVLAD pooling",
      "source_text":"POOLING STRATEGIES ::: NetVLAD pooling\nThe NetVLAD pooling strategy was initially developed for place recognition ..."}
   ]}
]}
\end{Verbatim}
LongBench prepends the document-content instruction to this JSON. The label
\texttt{level2.1} denotes a child of \texttt{level2}, not a new relevance rank.
LaRA instead formats the retained source text as numbered chunks and attaches
ancestor-title provenance where available. This format change does not change
which nodes were selected.
\clearpage

\section{Experimental Configurations and Evaluation}
\label{app:protocol-audit}
\subsection{Benchmarks and Sample Alignment}
\label{app:sample-alignment}
For each dataset, all evaluated methods, all four answer models, and all three
answer-generation runs use the same question set. LongBench examples are aligned
by question and reference answers, and LaRA examples by task, context type,
question, and reference answer. Each method--model score is averaged over the
same questions and three runs. Overall is the equally weighted mean of the four
model scores, so every method is evaluated on the same question--model
combinations.

\subsection{Answer Models and Judge Configurations}
\label{app:answer-configurations}
LongBench conditions use the benchmark's task-specific answer templates.
LaRA uses full-document prompts for Raw and retrieved-context prompts for
compressed inputs. Frozen context files are reused across answer runs.
LaRA uses a fixed gpt-oss-20b judge.

\subsection{Baseline Implementations and Context Budgets}
\label{app:baseline-configurations}
Table~\ref{tab:baseline-settings} summarizes the context-construction settings,
which are separate from the answer-model configuration. The additional
comparisons adjust compression budgets or pruning thresholds toward the retained
length of \ours. These targets do not impose identical per-question input sizes.

\begin{table}[htbp]
\centering
\small
\caption{Models and context settings for the main baseline comparison.}
\label{tab:baseline-settings}
\begin{tabularx}{\textwidth}{@{}lp{0.27\textwidth}>{\raggedright\arraybackslash}X@{}}
\toprule
Method & Compression or retrieval model & Configuration \\
\midrule
LongLL & Llama-2-7B & Requested retention rate 0.25. \\
Provence & DeBERTa-v3 reranker & Sentence threshold 0.1, title retention enabled,
no reordering. \\
HiChunk & Youtu-HiChunk and BGE-M3 & 200-token chunks, maximum depth 10,
top-128 retrieval and Auto-Merge, 4,096-token context budget. \\
BRIEF-Pro & BRIEF-Pro checkpoint & 20-sentence instruction, 1,024-token
output limit. \\
LongRefiner & Qwen2.5-3B-Instruct & Query analysis,
document structuring, and global selection, with BGE reranking and a 2,048-token
target. \\
\bottomrule
\end{tabularx}
\end{table}

For the adjusted comparisons, LongLL targets retention rates of 0.60/0.30/0.30
on Qasper/HotpotQA/2WikiMQA and 0.60 on LaRA. Provence uses threshold 0.001
on LongBench and 0.0001 on LaRA. HiChunk changes its context budget, while
BRIEF-Pro uses a 240-sentence instruction and a 16,384-token output limit.
LongRefiner targets the corresponding mean character ratios of \ours through
its adaptation layer. Actual retained lengths are reported in
Table~\ref{tab:length-adjusted-results}.

\clearpage

\section{Supplementary Quantitative Results}
\label{app:supplementary-results}
\subsection{Main-Comparison Details and Run-Level Intervals}
\label{app:main-result-details}
Tables~\ref{tab:longbench-main} and~\ref{tab:lara-main} report the main
comparison. Scores are averaged over three answer runs. Overall first averages
models within a task, and the LongBench summary gives equal weight to its three
tasks. Table~\ref{tab:run-level-main} retains the
run-level means and intervals from the main comparison.
\begin{table*}[t]
\centering
\small
\caption{Run-level variability for the main comparison. Each entry is the
four-model Overall mean with its 95\% Student-$t$ interval over three decoding
runs. The intervals characterize run variability on the fixed evaluation sample
and contexts only.}
\label{tab:run-level-main}
\fontsize{8.2}{9.8}\selectfont
\setlength{\tabcolsep}{2.5pt}
\renewcommand{\arraystretch}{1.1}
\begin{tabular*}{\textwidth}{@{\extracolsep{\fill}}lrrrrrrr@{}}
\toprule
Dataset & Raw & LongLL & Provence & HiChunk & BRIEF-Pro & LongRefiner & \ours \\
\midrule
Qasper & \scoreci{49.26}{0.99} & \scoreci{27.15}{1.05} & \scoreci{47.17}{0.90} & \scoreci{48.92}{0.48} & \scoreci{48.37}{0.74} & \scoreci{30.50}{0.35} & \scoreci{50.25}{0.22} \\
HotpotQA & \scoreci{69.51}{0.72} & \scoreci{56.55}{1.08} & \scoreci{66.01}{0.78} & \scoreci{66.62}{2.41} & \scoreci{68.18}{0.51} & \scoreci{57.64}{1.87} & \scoreci{71.23}{1.31} \\
2WikiMQA & \scoreci{70.63}{0.73} & \scoreci{56.54}{0.87} & \scoreci{61.82}{0.70} & \scoreci{69.69}{0.23} & \scoreci{69.34}{1.16} & \scoreci{45.80}{0.45} & \scoreci{73.09}{0.09} \\
LaRA & \scoreci{80.17}{0.30} & \scoreci{63.20}{0.29} & \scoreci{72.88}{0.28} & \scoreci{73.96}{1.14} & \scoreci{72.37}{1.19} & \scoreci{40.89}{1.48} & \scoreci{80.59}{0.37} \\
\bottomrule
\end{tabular*}
\end{table*}

\subsection{Compression--Performance Trade-off}
\label{app:compression-tradeoff}
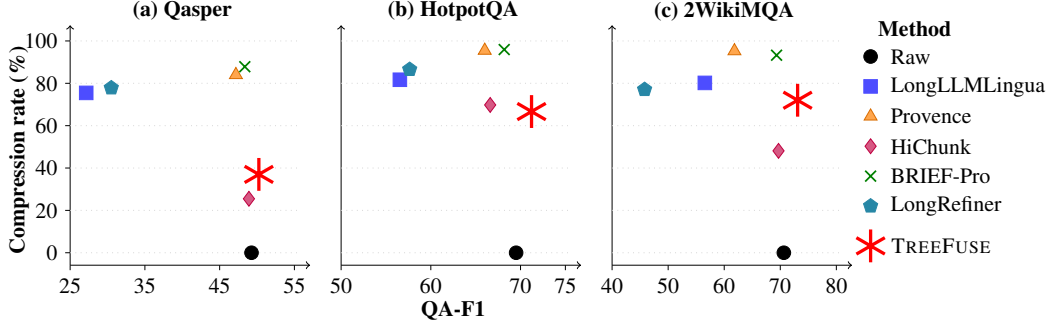
\begin{figure*}[t]
\centering
\resizebox{\textwidth}{!}{%
\begin{tikzpicture}
  \begin{scope}[xshift=0cm]
    \draw[->] (0,0) -- (3.85,0);
    \draw[->] (0,0) -- (0,3.85);
    \foreach \x/\lab in {0/25,1.2/35,2.4/45,3.6/55} {\draw (\x,0)--(\x,-0.08) node[below] {\lab};}
    \foreach \y/\lab in {0.2/0,0.88/20,1.56/40,2.24/60,2.92/80,3.60/100} {\draw (0,\y)--(-0.08,\y) node[left] {\lab}; \draw[dotted,gray!35] (0,\y)--(3.65,\y);}
    \node[font=\normalsize\bfseries] at (1.8,4.08) {(a) Qasper};
    \draw[only marks,mark=*,mark options={black,fill=black,mark size=3.1pt}] plot coordinates {(2.9116,0.2000)};
    \draw[only marks,mark=square*,mark options={blue!70,fill=blue!70,mark size=3.1pt}] plot coordinates {(0.2579,2.7662)};
    \draw[only marks,mark=triangle*,mark options={orange!85!black,fill=orange!70,mark size=3.4pt}] plot coordinates {(2.6609,3.0555)};
    \draw[only marks,mark=diamond*,mark options={purple,fill=purple!65,mark size=3.4pt}] plot coordinates {(2.8701,1.0661)};
    \draw[only marks,mark=x,mark options={green!50!black,thick,mark size=3.5pt}] plot coordinates {(2.8039,3.1866)};
    \draw[only marks,mark=pentagon*,mark options={cyan!60!black,fill=cyan!60!black,mark size=3.4pt}] plot coordinates {(0.6605,2.8462)};
    \draw[only marks,mark=asterisk,mark options={red,line width=1.5pt,mark size=7.5pt}] plot coordinates {(3.0295,1.4578)};
  \end{scope}
  \begin{scope}[xshift=4.35cm]
    \draw[->] (0,0) -- (3.85,0);
    \draw[->] (0,0) -- (0,3.85);
    \foreach \x/\lab in {0/50,1.44/60,2.88/70,3.6/75} {\draw (\x,0)--(\x,-0.08) node[below] {\lab};}
    \foreach \y in {0.2,0.88,1.56,2.24,2.92,3.60} {\draw[dotted,gray!35] (0,\y)--(3.65,\y);}
    \node[font=\normalsize\bfseries] at (1.8,4.08) {(b) HotpotQA};
    \draw[only marks,mark=*,mark options={black,fill=black,mark size=3.1pt}] plot coordinates {(2.8094,0.2000)};
    \draw[only marks,mark=square*,mark options={blue!70,fill=blue!70,mark size=3.1pt}] plot coordinates {(0.9431,2.9771)};
    \draw[only marks,mark=triangle*,mark options={orange!85!black,fill=orange!70,mark size=3.4pt}] plot coordinates {(2.3056,3.4461)};
    \draw[only marks,mark=diamond*,mark options={purple,fill=purple!65,mark size=3.4pt}] plot coordinates {(2.3929,2.5724)};
    \draw[only marks,mark=x,mark options={green!50!black,thick,mark size=3.5pt}] plot coordinates {(2.6182,3.4620)};
    \draw[only marks,mark=pentagon*,mark options={cyan!60!black,fill=cyan!60!black,mark size=3.4pt}] plot coordinates {(1.1007,3.1422)};
    \draw[only marks,mark=asterisk,mark options={red,line width=1.5pt,mark size=7.5pt}] plot coordinates {(3.0569,2.4666)};
  \end{scope}
  \begin{scope}[xshift=8.70cm]
    \draw[->] (0,0) -- (3.85,0);
    \draw[->] (0,0) -- (0,3.85);
    \foreach \x/\lab in {0/40,0.9/50,1.8/60,2.7/70,3.6/80} {\draw (\x,0)--(\x,-0.08) node[below] {\lab};}
    \foreach \y in {0.2,0.88,1.56,2.24,2.92,3.60} {\draw[dotted,gray!35] (0,\y)--(3.65,\y);}
    \node[font=\normalsize\bfseries] at (1.8,4.08) {(c) 2WikiMQA};
    \draw[only marks,mark=*,mark options={black,fill=black,mark size=3.1pt}] plot coordinates {(2.7569,0.2000)};
    \draw[only marks,mark=square*,mark options={blue!70,fill=blue!70,mark size=3.1pt}] plot coordinates {(1.4889,2.9265)};
    \draw[only marks,mark=triangle*,mark options={orange!85!black,fill=orange!70,mark size=3.4pt}] plot coordinates {(1.9642,3.4382)};
    \draw[only marks,mark=diamond*,mark options={purple,fill=purple!65,mark size=3.4pt}] plot coordinates {(2.6718,1.8353)};
    \draw[only marks,mark=x,mark options={green!50!black,thick,mark size=3.5pt}] plot coordinates {(2.6405,3.3711)};
    \draw[only marks,mark=pentagon*,mark options={cyan!60!black,fill=cyan!60!black,mark size=3.4pt}] plot coordinates {(0.5218,2.8201)};
    \draw[only marks,mark=asterisk,mark options={red,line width=1.5pt,mark size=7.5pt}] plot coordinates {(2.9783,2.6482)};
  \end{scope}
  \node[rotate=90,font=\normalsize\bfseries] at (-0.82,1.85) {Compression rate (\%)};
  \node[font=\normalsize\bfseries] at (6.15,-0.72) {QA-F1};
  \begin{scope}[xshift=12.85cm,yshift=3.35cm]
    \node[anchor=west,font=\normalsize\bfseries] at (0,0.45) {Method};
    \draw[only marks,mark=*,mark options={black,fill=black,mark size=3.1pt}] plot coordinates {(0,0)}; \node[anchor=west] at (0.20,0) {Raw};
    \draw[only marks,mark=square*,mark options={blue!70,fill=blue!70,mark size=3.1pt}] plot coordinates {(0,-0.48)}; \node[anchor=west] at (0.20,-0.48) {LongLLMLingua};
    \draw[only marks,mark=triangle*,mark options={orange!85!black,fill=orange!70,mark size=3.4pt}] plot coordinates {(0,-0.96)}; \node[anchor=west] at (0.20,-0.96) {Provence};
    \draw[only marks,mark=diamond*,mark options={purple,fill=purple!65,mark size=3.4pt}] plot coordinates {(0,-1.44)}; \node[anchor=west] at (0.20,-1.44) {HiChunk};
    \draw[only marks,mark=x,mark options={green!50!black,thick,mark size=3.5pt}] plot coordinates {(0,-1.92)}; \node[anchor=west] at (0.20,-1.92) {BRIEF-Pro};
    \draw[only marks,mark=pentagon*,mark options={cyan!60!black,fill=cyan!60!black,mark size=3.4pt}] plot coordinates {(0,-2.40)}; \node[anchor=west,align=left] at (0.20,-2.40) {LongRefiner};
    \draw[only marks,mark=asterisk,mark options={red,line width=1.5pt,mark size=7.5pt}] plot coordinates {(0,-3.04)}; \node[anchor=west] at (0.20,-3.04) {\ours};
  \end{scope}
\end{tikzpicture}
}
\caption{Compression--performance trade-off on Qasper, HotpotQA, and
2WikiMQA, averaged over Qwen3.8-Max, DeepSeek-V4-Pro, Kimi-K2.6, and GLM-5.3.
Compression rate is the percentage of raw context characters removed. The
vertical axis starts below 0\% so Raw remains visible. Moving right indicates
better QA-F1, whereas moving upward indicates stronger compression.}
\label{fig:longbench-tradeoff}
\end{figure*}

Figure~\ref{fig:longbench-tradeoff} relates four-model average QA-F1 to the
proportion of context removed on each LongBench task. We define compression rate
as $100\times(1-|C_m|/|C_{\mathrm{raw}}|)$, using mean context character counts
on the aligned examples. Points farther right indicate better answering, while
points higher in the plot indicate greater compression. \ours achieves the
highest average QA-F1 in all three panels while removing 58.55\% of the context
on average. More aggressive compression, as in BRIEF-Pro, produces shorter
inputs but lower average QA-F1. Additional comparisons in
Appendix~\ref{app:length-adjusted-results} adjust baseline budgets toward
\ours's average retained-context length. \ours retains a positive LongBench-average
margin over the strongest external baseline for each answer model in these
comparisons. Appendix~\ref{app:cost-benefit} extends the analysis to
successful-path input-side work, including amortized tree construction and
query-selection inputs.

\subsection{Comparisons with Adjusted Context Lengths}
\label{app:length-adjusted-results}
Table~\ref{tab:length-adjusted-results} adjusts baseline budgets toward the
retained length of \ours and reports the resulting character ratios. These are
not exact per-query budget matches. Within these comparisons, all methods and
all four answer models use the same question set for each dataset across three
answer runs, following the alignment procedure in
Appendix~\ref{app:sample-alignment}. The generation configurations follow
Appendix~\ref{app:answer-configurations}.

\begin{table*}[t]
\centering
\small
\caption{Additional comparisons with adjusted baseline context lengths.
LongBench reports mean QA-F1 across the three tasks and LaRA reports accuracy.
Context is the mean percentage of raw-context characters retained across the
corresponding models and tasks.}
\label{tab:length-adjusted-results}
\begin{tabular*}{\textwidth}{@{\extracolsep{\fill}}llrrrrr@{}}
\toprule
Benchmark & Method & Qwen & DeepSeek & Kimi & GLM & Context (\%) \\
\midrule
\multirow{7}{*}{LB}
 & Raw & 69.59 & 67.56 & 52.42 & 63.02 & 100.00 \\
 & LongLLMLingua & 63.11 & 55.90 & 42.69 & 53.21 & 38.76 \\
 & Provence & 69.43 & 67.32 & 53.10 & 63.45 & 35.96 \\
 & HiChunk & 66.54 & 64.62 & 50.27 & 61.11 & 38.96 \\
 & BRIEF-Pro & 68.71 & 63.84 & 54.28 & 62.29 & 23.75 \\
 & LongRefiner & 55.43 & 41.95 & 42.02 & 45.11 & 30.89 \\
 & \ours & \textbf{71.07} & \textbf{68.00} & \textbf{56.19} & \textbf{64.21} & 41.46 \\
\midrule
\multirow{7}{*}{LaRA}
 & Raw & 79.23 & 76.55 & 80.70 & \textbf{84.73} & 100.00 \\
 & LongLLMLingua & 75.09 & 72.04 & 74.69 & 78.85 & 55.64 \\
 & Provence & 78.74 & 78.05 & 79.87 & 84.57 & 47.61 \\
 & HiChunk & 77.80 & 76.14 & 78.93 & 82.05 & 51.46 \\
 & BRIEF-Pro & 76.19 & 74.33 & 76.96 & 80.96 & 16.94 \\
 & LongRefiner & 44.92 & 41.75 & 42.57 & 45.31 & 19.88 \\
 & \ours & \textbf{79.35} & \textbf{78.17} & \textbf{81.01} & 84.04 & 56.13 \\
\bottomrule
\end{tabular*}
\end{table*}

\subsection{Context Savings and Construction Amortization}
\label{app:cost-benefit}
Reducing the answer context also introduces tree-construction and query-selection
work. We quantify this trade-off using the archived successful-path input-token
accounting, denoted P1. This model counts the inputs to one successful tree
construction per document, one relevance-scoring pass and three cutoff requests
per query, and the final answer request. Local BM25 selection adds no LLM request.
We use input-side work because context construction directly changes how many
tokens each stage must read while holding the question, answer model, answer
instruction, decoding configuration, and generation budget fixed. P1 measures
whether amortized preprocessing is covered by reduced reader prefill under an
explicit reader-to-preprocessor weight.

\paragraph{Preprocessing and reuse.}
Let $D$ be the number of indexed documents and $N$ the number of associated
questions. With $Q=N/D$ questions per document, the mean preprocessing input per
question is
\[
C_{\mathrm{pre}}=\frac{T_{1,\mathrm{in}}}{N}
                 +\frac{T_{2,\mathrm{in}}}{N}
=\frac{T_{1,\mathrm{in}}/D}{Q}+\frac{T_{2,\mathrm{in}}}{N},
\]
where $T_{1,\mathrm{in}}$ and $T_{2,\mathrm{in}}$ are the total successful-path
construction and selection inputs. Table~\ref{tab:context-benefit} reports their
amortized contributions. LaRA shares 64 document trees across 887 questions,
reducing construction input to 2,209 tokens per question despite its longer
documents. This reuse affects construction, while query-conditioned selection
is charged separately for every question.

\begin{table}[htbp]
\centering
\small
\caption{Preprocessing input-token accounting. Construction and selection
columns are per-question Qwen3.6 inputs. $D$ and $N$ describe the indexed
collection used to amortize preprocessing. Values are rounded.}
\label{tab:context-benefit}
\setlength{\tabcolsep}{4pt}
\begin{tabular*}{\textwidth}{@{\extracolsep{\fill}}lrrrrrr@{}}
\toprule
Dataset & $D$ & $N$ & $N/D$ & Construction & Selection & $C_{\mathrm{pre}}$ \\
\midrule
Qasper & 148 & 200 & 1.351 & 4,865 & 14,318 & 19,183 \\
HotpotQA & 200 & 200 & 1.000 & 14,827 & 33,418 & 48,245 \\
2WikiMQA & 200 & 200 & 1.000 & 8,903 & 25,403 & 34,306 \\
LaRA & 64 & 887 & 13.859 & 2,209 & 19,485 & 21,694 \\
\bottomrule
\end{tabular*}
\end{table}

\paragraph{Model-weighted input work.}
Let $T_{\mathrm{raw}}$ and $T_{\mathrm{sel}}$ be the answer model's mean input
tokens per question with full and selected contexts. We use $\kappa$ for its
relative input-scale weight, distinct from the serialization rule $\rho$ in
Appendix~\ref{app:selection-implementation} (Algorithm~\ref{alg:selection}).
The comparison is
\[
C_{\mathrm{raw}}^{\mathrm{P1}}=\kappa T_{\mathrm{raw}},\qquad
C_{\mathrm{ours}}^{\mathrm{P1}}=C_{\mathrm{pre}}+\kappa T_{\mathrm{sel}}.
\]
For $T_{\mathrm{raw}}>T_{\mathrm{sel}}$, the break-even weight is
\[
\boxed{\kappa^*=\frac{C_{\mathrm{pre}}}
{T_{\mathrm{raw}}-T_{\mathrm{sel}}}}.
\]
This threshold is the primary input-work result: it states how expensive one reader input
token must be, relative to one Qwen3.6 preprocessing input token, for the reader
context reduction to amortize construction and selection. A deployment can compare
the threshold directly with its own reader-to-preprocessor ratio.

For a transparent secondary instantiation, we set $\kappa$ to activated-parameter
ratios obtained from public documentation and model configurations: 11.089 for
Kimi-K2.6, 17.149 for DeepSeek-V4-Pro, 31.502 for Qwen3.8-Max, and 14.034 for
GLM-5.3. The corresponding activated-parameter counts are 31.686B, 49B, 90.014B,
and 40.101B, respectively, against the 2.857B Qwen3.6 reference. 
\begin{table*}[t]
\centering
\small
\caption{Break-even thresholds and modeled reductions in proxy-weighted
successful-path input work under the stated $\kappa$. Raw and \ours are Qwen3.6
input-token equivalents per question, including preprocessing for \ours. Percentages are computed from the displayed rounded proxy totals.}
\label{tab:input-scale-cost}
\setlength{\tabcolsep}{3pt}
\begin{tabular*}{\textwidth}{@{\extracolsep{\fill}}llrrrr@{}}
\toprule
Answer model & Dataset & $\kappa^*$ & Raw & \ours & Modeled red. (\%) \\
\midrule
\multirow{4}{*}{\shortstack[l]{Qwen3.8-Max\\$\kappa=31.502$}}
 & Qasper & 10.24 & 163,591 & 123,773 & 24.3 \\
 & HotpotQA & 5.40 & 424,765 & 191,750 & 54.9 \\
 & 2WikiMQA & 6.36 & 238,521 & 102,885 & 56.9 \\
 & LaRA & 1.66 & 923,552 & 532,473 & 42.3 \\
\midrule
\multirow{4}{*}{\shortstack[l]{DeepSeek-V4-Pro\\$\kappa=17.149$}}
 & Qasper & 10.64 & 87,571 & 75,833 & 13.4 \\
 & HotpotQA & 5.75 & 219,470 & 123,776 & 43.6 \\
 & 2WikiMQA & 6.83 & 122,023 & 70,221 & 42.5 \\
 & LaRA & 1.82 & 457,116 & 273,947 & 40.1 \\
\midrule
\multirow{4}{*}{\shortstack[l]{Kimi-K2.6\\$\kappa=11.089$}}
 & Qasper & 10.51 & 56,804 & 55,741 & 1.9 \\
 & HotpotQA & 5.71 & 142,578 & 97,075 & 31.9 \\
 & 2WikiMQA & 6.74 & 79,792 & 57,694 & 27.7 \\
 & LaRA & 1.76 & 305,651 & 190,778 & 37.6 \\
\midrule
\multirow{4}{*}{\shortstack[l]{GLM-5.3\\$\kappa=14.034$}}
 & Qasper & 10.59 & 71,650 & 65,416 & 8.7 \\
 & HotpotQA & 5.73 & 180,233 & 110,233 & 38.8 \\
 & 2WikiMQA & 6.75 & 101,034 & 64,033 & 36.6 \\
 & LaRA & 1.77 & 385,194 & 234,980 & 39.0 \\
\bottomrule
\end{tabular*}
\end{table*}

\paragraph{Sensitivity to the reader-weight proxy.}
For a candidate weight $\kappa$, the relative modeled reduction in
proxy-weighted successful-path input work is
\[
s(\kappa)=\frac{\kappa T_{\mathrm{raw}}-
\left(C_{\mathrm{pre}}+\kappa T_{\mathrm{sel}}\right)}
{\kappa T_{\mathrm{raw}}}.
\]
Thus $s(\kappa^*)=0$, and the sign changes exactly at the break-even threshold.
Table~\ref{tab:input-scale-cost} reports $\kappa^*$ directly, allowing readers to
substitute a different measured or assumed reader-to-preprocessor weight without
treating the activated-parameter proxy as exact. Because answer-model tokenizers
produce slightly different token counts, each task has a narrow range of
model-specific thresholds. The closest case is Qasper with Kimi-K2.6:
$\kappa=11.089$ is only 5.2\% above $\kappa^*=10.51$, so its 1.9\% modeled margin
is sensitive to modest proxy misspecification. The other task-level threshold
ranges are 5.40--5.75 for HotpotQA, 6.36--6.83 for 2WikiMQA, and 1.66--1.82 for
LaRA.

\paragraph{Quantitative implications.}
The break-even thresholds are 5.40--6.83 on HotpotQA and 2WikiMQA,
10.24--10.64 on Qasper, and 1.66--1.82 on LaRA. The low LaRA thresholds reflect
reuse of 64 document trees across 887 questions. Under the stated
activated-parameter proxies, all 16 combinations have positive modeled
reductions in proxy-weighted successful-path input work, ranging from 1.9\% to
56.9\% (Table~\ref{tab:input-scale-cost}). For example, the modeled reductions
are 54.9\% and 56.9\% for Qwen3.8-Max on HotpotQA and 2WikiMQA, respectively.

The smallest modeled margin is Qasper with Kimi-K2.6: $\kappa=11.089$ is only
5.2\% above $\kappa^*=10.51$, producing a 1.9\% modeled reduction. Its sign is
therefore sensitive to modest misspecification of the activated-parameter proxy.
In contrast, LaRA's proxy weights are far above its 1.66--1.82 break-even range.
For Qwen3.8-Max on LaRA, the stated proxy corresponds to 391,079 reference
input-token equivalents per question after amortized preprocessing.

\paragraph{Accounting scope.}
Construction and selection inputs use the Qwen3.6 tokenizer. Answer inputs for
all four answer models are re-tokenized with their corresponding tokenizers.
Answer-input averages use the common questions across three Raw and three \ours
runs. Preprocessing is amortized over the indexed collection in
Table~\ref{tab:context-benefit}.

\clearpage
\section{Source-Evidence Analysis}
\label{app:evidence-metrics}
\subsection{Evidence Annotations and Source Alignment}
\label{app:gold-alignment}

Qasper uses mapped textual evidence with alternative annotation sets.
HotpotQA and 2WikiMQA use mapped supporting passages. The component results are
reported in Table~\ref{tab:evidence-components-main}; the metrics quantify
annotated source preservation rather than semantic answer sufficiency.

\subsection{Evidence Metric Definitions}
\label{app:evidence-definitions}

\paragraph{Complete evidence coverage.}
Let $\mathcal{G}_i$ contain the eligible evidence sets for question $i$ and
$r(u,C_i)$ denote normalized literal coverage of evidence unit $u$ in context
$C_i$. The matcher checks normalized source text and sentence or line segments.
We define
\begin{equation}
A_i=\mathbb{1}\!\left[\exists G\in\mathcal{G}_i:
       r(u,C_i)\geq0.999\ \text{for every }u\in G\right],
\qquad
\mathrm{All\text{-}hit}=\frac{100}{N}\sum_{i=1}^{N}A_i.
\label{eq:all-hit}
\end{equation}
Thus, one completely covered alternative is sufficient. Passage-level All-hit
measures literal preservation, not the validity of a reasoning chain or semantic
answer sufficiency.

\paragraph{Evidence concentration.}
For purity, selected content is aligned to normalized lexical tokens in the
original document. Let $P_i$ be the set of distinct recovered source-token
positions, $G_i^*$ the mapped reference-token set, and $U_i$ the number of
unmapped output tokens after method-specific content extraction. Then
\begin{equation}
\mathrm{Evidence\ Purity}
=\frac{100}{N}\sum_{i=1}^{N}
 \frac{|P_i\cap G_i^*|}{\max(1,|P_i|+U_i)}.
\label{eq:evidence-purity}
\end{equation}
For Qasper, eligible textual alternatives must map to source tokens, and the
scorer chooses the alternative with greatest token recall, breaking ties by
source-token intersection-over-union. This rule differs from the literal
complete-coverage test used for All-hit. For the multi-hop tasks, reference
positions are obtained from the mapped supporting-passage spans. Whole units
are matched first, with line-level fallback. Token-pruned outputs use sequence
alignment. Structural wrappers are excluded where the parser extracts source
content. Recovered source positions are deduplicated, so the denominator is not
a literal count of every token sent to the answer model. Unmapped text can
include valid paraphrases, and unannotated context can still support answering.
Purity is consequently an annotation-based concentration proxy rather than the
fraction of all useful information in the output.

\subsection{Equal-Weight ECP Rationale}
\label{app:ecp-uncertainty}
To summarize completeness and concentration without introducing a tunable
trade-off, we use their equal-weight arithmetic mean:
\begin{equation}
\mathrm{ECP}=\frac{1}{2}\left(\mathrm{All\text{-}hit}
             +\mathrm{Evidence\ Purity}\right).
\label{eq:ecp}
\end{equation}
All-hit measures whether a complete annotated evidence set is retained, whereas
Evidence Purity measures how concentrated the selected content is around mapped
evidence. Equal weighting treats these as complementary diagnostic dimensions
and avoids choosing a dataset-dependent preference between completeness and
concentration. Because both components are percentages, ECP remains on the same
$0$--$100$ scale. The same aggregation is applied to every method and dataset;
it is a source-evidence diagnostic rather than a model of answer correctness.

\clearpage

\section{Ablation Details and Case Studies}
\label{app:ablations}
\subsection{Selection, Representation, and Construction Variants}
\label{app:ablation-variants}
Tables~\ref{tab:ablation-selection}--\ref{tab:ablation-construction} use
Qwen3.8-Max and a common question set within each comparison over three answer
runs. Table~\ref{tab:ablation-interfaces}
summarizes the changes to the full method. BM denotes approximate character-budget
matching, not exact per-query token equality.

\begin{table}[htbp]
\centering
\small
\caption{Changes made by each ablation variant.}
\label{tab:ablation-interfaces}
\begin{tabularx}{\textwidth}{@{}l>{\raggedright\arraybackslash}X@{}}
\toprule
Variant & Change relative to the full method \\
\midrule
Semantic-only & Retain semantic selections and their ancestor context. \\
Lexical-only (BM) & Use lexical node selection with ancestor closure and an
approximately matched character budget. \\
Semantic-only (BM) & Add fixed-seed random nodes to the semantic selection to
approach the full method's context length. \\
Fixed 200 & Replace semantic units with 200-token chunks and apply the same
selection framework. \\
Chonkie & Replace units with the configured Chonkie semantic chunks. \\
Flat & Retain the semantic units and source text, but remove their hierarchy. \\
Multi-stage & Replace the constructor with the same-model multi-stage pipeline. \\
HiChunk adapter & Map HiChunk-derived units into the node format used by our
selector rather than using the complete HiChunk retrieval system. \\
\bottomrule
\end{tabularx}
\end{table}

\subsection{Tree-Construction Configurations}
\label{app:construction-variants}
Table~\ref{tab:ablation-construction} compares representations supplied to the
same downstream selection framework. The single-pass and multi-stage
constructors both use Qwen3.6-35B-A3B. Single-pass construction generates the
hierarchy and node descriptions jointly from the document, as described in
Appendix~\ref{app:construction-selection}. It denotes a joint generation design,
not a guarantee of exactly one request when output validation requires a retry.

\paragraph{Multi-stage construction.}
The alternative pipeline separates structure extraction, source anchoring, and
content enrichment. It first partitions the source into indexed chunks with an
800-token target while respecting passage boundaries, then forms groups of up
to 20,000 tokens with one-chunk overlap. An initial call proposes an outline
containing titles, hierarchy paths, and chunk indices. Subsequent groups extend
the existing outline, with an additional repair step for uncovered chunks when
needed. A combined validation and localization stage checks candidate headings
against their source chunks and identifies section boundaries. These anchors
are mapped to document character positions to assemble the tree and attach
local source text. Oversized leaf nodes can be subdivided before bottom-up
content enrichment. Leaf descriptions use their own text, whereas internal-node
descriptions combine local introductory text with child summaries. The pipeline
receives the complete document, but individual calls operate on groups, chunks,
or node material. The initial outline call can see the full document when it
fits within one group.

\paragraph{Configuration and downstream selection.}
Validation and localization batch up to five chunks per call. Leaf subdivision
uses a 16,000-token threshold and a recursion-depth limit of two. Content
enrichment produces summaries and facts, batching up to six nodes with
16,000 characters of material per batch and an 8,000-character per-node limit.
Both representations feed the semantic scorer, adaptive cutoff, ancestor
completion, and lexical-union procedure. The selector uses the P90 lexical
cutoff on both HotpotQA and 2WikiMQA. Node boundaries, metadata, and the selected
content can therefore differ despite the shared selection framework.

\paragraph{HiChunk representation adapter.}
This configuration adapts cached HiChunk-derived segments and hierarchy levels
to the common representation interface required by our downstream selector.
The adaptation is introduced only for compatibility with the shared selection
framework used in this comparison. It therefore evaluates a HiChunk-derived
representation with our downstream selection procedure, rather than the
complete HiChunk retrieval and Auto-Merge system.

\subsection{Conditional Evidence and Answer Analysis}
\label{app:conditional-ablation}
Table~\ref{tab:conditional-evidence} pairs each variant's saved context with
its Qwen3.8-Max answers. Span Recall measures length-weighted literal coverage
of annotated source units. Among alternative evidence sets, the scorer prefers
complete coverage, then the highest weighted coverage. QA-F1 is averaged over
three runs for each question. Each reported difference is Full minus the named
variant.

\begin{table*}[t]
\centering
\small
\setlength{\tabcolsep}{3pt}
\caption{Conditional evidence and answer differences with Qwen3.8-Max. Each
cell reports $\Delta$Span Recall in percentage points / $\Delta$QA-F1 in score
points. BM denotes approximate character-budget matching.}
\label{tab:conditional-evidence}
\begin{tabular*}{\textwidth}{@{\extracolsep{\fill}}lrrrr@{}}
\toprule
Full minus & Qasper & HotpotQA & 2WikiMQA & Pooled \\
\midrule
Semantic-only & +1.14 / +0.11 & +1.86 / +0.40 & +0.53 / +0.56 & +1.17 / +0.36 \\
Semantic-only (BM) & +1.14 / $-1.14$ & $-0.07$ / +2.42 & +0.22 / +2.76 & +0.41 / +1.42 \\
Lexical-only (BM) & +16.45 / +0.60 & +18.96 / +4.17 & +28.46 / +8.14 & +21.50 / +4.45 \\
\bottomrule
\end{tabular*}
\end{table*}

Semantic-only (BM) augments the original semantic selections with fixed-seed random
nodes to approximate the full method's context length. Lexical-only (BM) uses
lexical node selection with ancestor closure. The generation-setting differences
recorded in Appendix~\ref{app:protocol-audit} apply to these comparisons as well.
They describe the available configurations rather than isolating every change
from answer-generation effects.

\subsection{Semantic Support and Lexical Supplementation Cases}
\label{app:selection-cases}
The examples below are drawn from saved Qwen3.8-Max outputs and their paired
contexts. They illustrate distinct selection behaviors rather than estimating
population-level success rates. Span Recall and QA-F1 are percentages and
three-run means, respectively. Displayed source excerpts are verbatim.

\paragraph{Lexical supplementation: recovering data provenance.}
A Qasper question asks, ``How is the dataset of hashtags sourced?'' The source
identifies ``1,268 randomly selected tweets in the Stanford Sentiment Analysis
Dataset'' and ``all 12,594 unique English hashtags and their associated tweets
from the same Stanford dataset.'' Compared with Semantic-only (BM), Full raises
Span Recall from 0 to 100 and mean QA-F1 from 0 to 71.11. All three
Semantic-only (BM) answers are ``unanswerable'', whereas all three Full answers
identify the Stanford Sentiment Analysis Dataset. This illustrates the value of
retaining source-specific provenance in the selected context.

\paragraph{Semantic support: connecting a mission to its destination.}
A HotpotQA question asks, ``Gary L. Bennett was a part of the space missions
that have a primary destination of what celestial body?'' The source links
Bennett to the Ulysses mission and states that Ulysses ``was a robotic space
probe whose primary mission was to orbit the Sun and study it at all latitudes.''
Compared with Lexical-only (BM), Full raises Span Recall from 2.10 to 82.89.
All three lexical-only answers name Jupiter, whereas all three Full answers name
the Sun, changing mean QA-F1 from 0 to 100. The example illustrates how the
selected source context can connect a named participant to the mission target.

\paragraph{Evidence enrichment without an answer-score change.}
Another HotpotQA question asks, ``Which California band, whose debut album
Adrenaline appeared in 1995, has been referred to as `the Radiohead of metal'?''
The source identifies Deftones and states, ``The band's experimental nature has
led some critics to describe them as \"the Radiohead of metal\".''
lexical-only (BM) and Full both answer Deftones in all three runs. Full increases
Span Recall from 87.28 to 96.34 while mean QA-F1 remains 100. Thus, identical
correct answers can arise from contexts with different annotated-evidence
coverage. These are examples from the recorded configurations described in
Appendix~\ref{app:answer-configurations}, not isolated causal tests.

\Needspace{8\baselineskip}
\subsection{Semantic Adequacy versus Boundary Exactness}
\label{app:boundary-adequacy-case}
\paragraph{Routing metadata and source evidence.}
Generated boundaries, hierarchy, summaries, and facts can affect which source
regions are selected. The relevant distinction is between \emph{evidence
containment} and exact recovery of annotation boundaries: the tree organizes
source regions for retrieval rather than reproducing a unique gold segmentation.
Titles and structural paths provide navigation, while summaries and facts guide
selection rather than substitute for source evidence in the answer context.
The source matcher anchors nodes to document text; when both summary and facts
are absent, scoring uses a source-text prefix
(Appendix~\ref{app:tree-construction}). The following case examines whether a
node broader than individual evidence units can still retain their support.

\paragraph{A coherent region containing the annotated mechanism.}
The Qasper example \texttt{qasper-125-5724368043143d51-q151} asks,
\emph{``How are weights dynamically adjusted?''} One annotated alternative
evidence set contains four units, all mapped to node \texttt{n8}. These units
explain why easy-negative examples dominate training, the modification that
multiplies soft probability $p$ by $(1-p)$, how this factor changes during
training to down-weight easy examples, and its connection to focal loss.
The node contains the mechanism together with its motivation and interpretation,
rather than assigning a separate annotation-aligned boundary to each unit.
The four-unit set is the alternative used for the case metrics below, not the
union of all annotations for this question; matching follows
Appendices~\ref{app:gold-alignment} and~\ref{app:evidence-definitions}.

\paragraph{Retained evidence and downstream answers.}
Table~\ref{tab:boundary-adequacy-case} compares the saved Lexical-only (BM)
and Full contexts. Full retains \textbf{all four evidence units} in the source
region, raising Span Recall from 0 to 100 and mean QA-F1 from 17.57 to 35.33.
The Lexical-only context omits all four units. Thus, complete recovery of this
annotated support occurs without exact unit-level segmentation. The QA-F1 gain
accompanies improved evidence coverage; it does not imply a perfectly correct
answer or isolate boundary placement as its cause.

\begin{table}[htbp]
\centering
\small
\setlength{\tabcolsep}{3pt}
\caption{Evidence retention for the Qasper weight-adjustment case. Span Recall
and QA-F1 use a 0--100 scale; QA-F1 averages three Qwen3.8-Max answer runs.
BM matches mean character budgets, not individual context lengths.}
\label{tab:boundary-adequacy-case}
\begin{tabular*}{\textwidth}{@{\extracolsep{\fill}}lrrrrr@{}}
\toprule
Condition & \shortstack{Evidence\\units} & \shortstack{Span\\Recall} & All-hit &
\shortstack{Mean\\QA-F1} & \shortstack{Context\\chars.} \\
\midrule
Lexical-only (BM) & 0/4 & 0.00 & No & 17.57 & 5,050 \\
Full (\ours) & 4/4 & 100.00 & Yes & 35.33 & 12,552 \\
\bottomrule
\end{tabular*}
\end{table}

\paragraph{Implication for construction reliability.}
This case establishes a narrower requirement than boundary-perfect
reconstruction: \textbf{a coherent source node can preserve the required
evidence without matching its annotation boundaries}. For source-grounded
selection to work, the relevant text must remain represented in selectable
nodes, those nodes must be retrieved, and their original text must reach the
reader. Boundary or metadata differences need not remove support when these
conditions remain satisfied. Conversely, source grounding does not prevent
incorrect metadata or structure from causing a selection miss. This is an
observed containment case, not a controlled corruption experiment or an estimate
of tree-construction error rates. It supports separating routing metadata from
answer evidence, rather than treating arbitrary construction errors as harmless.

\end{document}